\documentclass[]{article}
\usepackage{proceed2e}

\usepackage{times}
\usepackage[pdftex]{graphicx} 

\usepackage{qiangstyle}
\usepackage{times}

\title{Stein Variational Adaptive Importance Sampling}

\author{
{\bf Jun Han~~~~~~~~~~~~~~~~~~~~~~~~~~~~Qiang Liu} \\
Computer Science,
Dartmouth College,
Hanover, NH 03755 \\
\texttt{\{jun.han.gr, qiang.liu\}@dartmouth.edu}
}

\begin{document}

\maketitle


\begin{abstract}
We propose a novel adaptive importance sampling algorithm which incorporates Stein variational gradient decent algorithm~(SVGD) with importance sampling~(IS).
Our algorithm leverages the nonparametric transforms in SVGD to iteratively decrease the KL divergence between  importance proposals and target distributions.
The advantages of our algorithm are twofold:
1) it turns SVGD into a standard IS algorithm, allowing us to use
standard diagnostic and analytic tools of IS to evaluate and interpret the results, and 2)
it does not restrict the choice of the importance proposals to predefined distribution families like traditional (adaptive) IS methods. 
Empirical experiments demonstrate that our algorithm performs well on
evaluating partition functions of restricted Boltzmann machines and
testing likelihood of variational auto-encoders.
\end{abstract}

\section{INTRODUCTION}
Probabilistic modeling provides a fundamental framework for reasoning under uncertainty and modeling complex relations in machine learning. 
A critical challenge, however, is to develop efficient computational techniques for
approximating complex distributions. Specifically,
given a complex distribution $p(\bd x)$,
often known only up to a normalization constant,
we are interested estimating integral quantities $\E_{p}[f]$  for test functions $f.$
Popular approximation algorithms
include particle-based methods, such as
Monte Carlo, which construct a set of independent particles $\{\vv x_i\}_{i=1}^n$
whose empirical averaging $\frac{1}{n}\sum_{i=1}^n f(\vv x_i)$
forms unbiased estimates of $\E_p[f]$,
and 
variational inference (VI),
which approximates $p$ with a simpler surrogate distribution $q$
by minimizing a KL divergence objective function within a predefined parametric family of distributions.
Modern variational inference methods have found successful applications in
highly complex learning systems \citep[e.g., ][]{hoffman2013stochastic, kingma2013auto}. However, VI critically depends on the choice of parametric families and does not generally provide consistent estimators like particle-based methods.

Stein variational gradient descent (SVGD) is an alternative
framework that integrates both the particle-based and variational ideas.
It starts with a set of initial particles
$\{\vv x_i^0\}_{i=1}^n$, and
iteratively updates the particles using adaptively constructed {deterministic variable transforms:} 
$$\bd{x}_i^{\ell} \gets \T_{\ell}(\bd{x}_i^{\ell-1}), ~~~~\forall i= 1,\ldots, n,$$
where $\T_\ell$ is a variable transformation at the $\ell$-th iteration that maps old particles to new ones,
constructed adaptively at each iteration based on the most recent particles $\{\bd{x}_i^{\ell-1}\}_{i=1}^n$
that guarantee to push the particles ``closer'' to the target distribution $p$, in the sense that the KL divergence between the distribution of the particles and the target distribution $p$ can be iteratively decreased.
More details on the construction of $\T_\ell$ can be found in Section 2.

In the view of measure transport, SVGD iteratively transports the initial probability mass of the particles to the target distribution.
SVGD constructs a path of distributions that bridges the initial distribution $q_0$ to the target distribution $p$,
\begin{align}\label{equ:qt1}
q_{\ell} =  (\T_{\ell} \circ \cdots \circ \T_{1})\sharp q_0, \quad \ell=1, \ldots, K.
\end{align}
where $\T \sharp q$ denotes the push-forward measure of $q$ through the transform $\T$, that is the distribution of $\vv z = \T(\vv x)$ when $\vv x\sim q$.

The story, however, is complicated
by the fact that the transform $\T_\ell$ is practically constructed on the fly \emph{depending} on the recent particles $\{\vx_i^{\ell-1}\}_{i=1}^n$,
which introduces complex dependency between the particles at the next iteration,
whose theoretical understanding requires mathematical tools in interacting particle systems \citep[e.g.,][]{braun1977vlasov, spohn2012large, del2013mean} and propagation of chaos \citep[e.g.,][]{sznitman1991topics}.
As a result, $\{\vv x_i^{\ell}\}_{i=1}^n$ can not be viewed as i.i.d. samples from $q_\ell$.
This makes it
difficult to analyze the results of SVGD and quantify their bias and variance.

In this paper,
we propose a simple modification of SVGD
that ``decouples'' the particle interaction
and returns particles i.i.d. drawn from $q_\ell$;
we also develop a method to iteratively keep track of the importance weights of these particles,
which makes it possible to give consistent, or unbiased estimators within finite number of iterations of SVGD. 

Our method integrates 
 SVGD with importance sampling (IS) and combines their advantages:
it leverages the SVGD dynamics to obtain high quality proposals $q_\ell$ for IS
and also turns SVGD into a standard IS algorithm, inheriting the interpretability and theoretical properties of IS.
Another advantage of our proposed method is that it
provides an SVGD-based approach for estimating intractable normalization constants,
an inference problem that the original SVGD does not offer to solve.

\paragraph{Related Work}
Our method effectively turns SVGD into a nonparametric, adaptive importance sampling (IS) algorithm,
where the importance proposal $q_\ell$ is adaptively improved by the optimal transforms $\T_\ell$ which maximally decreases the KL divergence between the iterative distribution and the target distribution in a function space.
This is in contrast to
the traditional adaptive importance sampling methods
\citep[e.g.,][]{cappe2008adaptive, ryu2014adaptive, cotter2015parallel},
which optimize the proposal distribution from predefined distribution families $\{q_{\bd{\theta}}(\bd{x})\}$, 
often mixture families or exponential families. 
The parametric assumptions restrict the choice of the proposal distributions
and may give poor results when the assumption is inconsistent with
the target distribution $p$.  
The proposals $q_\ell$ in our method, however, are obtained by recursive variable transforms constructed in a nonparametric fashion
and become more complex as more transforms $\T_\ell$ are applied.
In fact, one can view $q_\ell$ as the result of pushing $q_0$ through a neural network with $\ell$-layers,
constructed in a non-parametric, layer-by-layer fashion,
which provides a much more flexible distribution family than typical parametric families such as mixtures or exponential families.

There has been a collection of recent works 
 \citep[such as][]{rezende2015variational, kingma2016improved, marzouk2016introduction, spantini2017inference},
that approximate the target distributions
with complex proposals obtained by iterative variable transforms in a similar way to our proposals $q_\ell$ in \eqref{equ:qt1}.
The key difference, however, is that these methods explicitly parameterize the transforms $\T_\ell$
and optimize the parameters by back-propagation,
while our method, by leveraging the nonparametric nature of SVGD,
constructs the transforms $\T_\ell$ sequentially in closed forms,
requiring no back-propagation.

The idea of constructing a path of distributions $\{q_\ell\}$ to bridge
the target distribution $p$ with a simpler distribution $q_0$
invites connection to ideas such as annealed importance sampling (AIS) \citep{neal2001annealed}
and path sampling (PS) \citep{gelman1998simulating}.  
These methods typically construct an annealing path using geometric averaging of the initial and target densities instead of variable transforms,
which does not build in a notion of variational optimization as the SVGD path.
In addition, it is often intractable to directly sample distributions on the geometry averaging path,
and hence  AIS and PS  need additional mechanisms in order to construct proper estimators.

{\bf Outlines} The reminder of this paper is organized as follows. 
Section 2 discusses Stein discrepancy and SVGD. 
We propose our main algorithm in Section 3, and a related method in Section 4. 
Section 5 provides empirical experiments and Section 6 concludes the paper. 

\section{STEIN VARIATIONAL GRADIENT DESCENT}
We introduce the basic idea of Stein variational gradient descent (SVGD) and Stein discrepancy. 
The readers are referred to \citet{liu2016stein} and \cite{liu2016kernelized} for more detailed introduction.

\paragraph{Preliminary}
We always assume $\vx=[x_1,\cdots, x_d]^\top \in \R^d$ in this paper.
Given a positive definite kernel $k(\vx,\vx')$, there exists an unique reproducing kernel Hilbert space (RKHS) $\H_0$,
formed by the closure of functions of form $f(\vx) = \sum_{i} a_i k(\vx,\vx_i)$ where $a_i \in \RR$, equipped with inner product
$\la f, ~ g\ra_{\H_0} = \sum_{ij}a_i k(\vx_i, \vx_j) b_j$ for $g(\vx) = \sum_j b_j k(\vx, \vx_j)$. 
Denote by $\H  = \H_0^d = \H_0 \times \cdots \times \H_0$ the vector-valued function space formed by $\vv f = [f_1, \ldots, f_d]^\top$, where $f_i \in \H_0$, $i=1,\ldots, d$, equipped with inner product $\la \vv f, ~ \vv g\ra_{\H}=\sum_{l=1}^d \la f_l, ~ g_l\ra_{\H_0}, $ for $\vv g =[g_1, \ldots, g_d]^\top.$ 
Equivalently, $\H$ is the closure of functions of form  $\vv f(\vx) = \sum_{i} \vv a_i k(\vx,\vx_i)$ where $\vv a_i \in \RR^d $ 
with inner product $\la \vv f, ~ \vv g\ra_{\H} = \sum_{ij}\vv a_i^\top \vv b_j k(\vx_i, \vx_j)$ for $\vv g(\vx) = \sum_{i} \vv b_i k(\vx,\vx_i)$. 
See e.g.,~\citet{berlinet2011reproducing} for more background on RKHS.  

\subsection{Stein Discrepancy as Gradient of KL Divergence}
Let $p(\vx)$ be a density function on $\R^d$ which we want to approximate. We assume that we know $p(\vx)$ only up to a normalization constant, that is,
\begin{equation}
\label{barp}
p(\vx) = \frac{1}{Z} \bar p(\vx), ~~~~~ Z = \int \bar p(\vx) d\vx,
\end{equation}
where we assume we can only calculate $\bar p(\vx)$ and $Z$ is a normalization constant (known as the partition function) that is intractable to calculate exactly.
We assume that $\log p(\vx)$ is differentiable w.r.t. $\vx$, and we have access to $\nabla \log p(\vx) = \nabla \log \bar p(\vx )$ which does not depend on $Z$.

The main idea of SVGD is to use a set of sequential deterministic transforms
to iteratively push a set of particles $\{\vx_i\}_{i=1}^n$ towards the target distribution:
 \begin{align}
 \label{update}
 \begin{split}
 & \vx_i \leftarrow  \T(\vx_i), ~~~~ \quad \forall i=1, 2,\cdots, n \\
& \T(\vx) =  \bd{x} +\epsilon \bd{\phi}(\bd{x}),
\end{split}
 \end{align}
 where we choose the transform $\T$ to be an additive perturbation by a velocity field  $\ff$,
 with a magnitude controlled by a step size
 $\epsilon$ that is assumed to be small. 

The key question is the choice of the velocity field $\ff $; this is done by choosing $\ff$ to maximally decrease the $\KL$ divergence between the distribution of particles and the target distribution. Assume the current particles are drawn from $q$, and $\T\sharp q$ is the distribution of the updated particles, that is, $\T\sharp q$ is the distribution of
$\bd{\vx}'=\T(\vx) = \bd{\vx}+\epsilon\bd{\phi}(\bd{\vx})$ when $\vx \sim q$.
The optimal $\ff$ should solve the following functional optimization:
 \begin{align}
 \label{vgddecrease}
 \begin{split}
\mathbb{D}(q ~||~ p)  \overset{def}{=}  \max_{\bd{\phi} \in \mathcal \F  \colon  ||\ff ||_{\F}\leq 1 } \bigg\{ - \frac{d}{d\epsilon}\KL( \T\sharp q \mid\mid p)~ \big|_{\epsilon=0}  \bigg\},
\end{split}
 \end{align}
where $\mathcal{F}$ is a vector-valued normed function space that contains the set of candidate velocity fields $\ff$. 

The maximum negative gradient value $\mathbb{D}(q ~||~ p)$ in \eqref{vgddecrease}
provides a discrepancy measure between two distributions $q$ and $p$ and is known as \emph{Stein discrepancy} \citep{gorham2015measuring, liu2016kernelized, chwialkowski2016kernel}: if $\mathcal{F}$ is taken to be large enough, we have $\mathbb{D}(q ~||~ p) = 0$ iff there exists no transform to further improve the KL divergence between $p$ and $q$, namely $p = q$.

It is necessary to use an infinite dimensional function space $\mathcal{F}$ to
obtain good transforms,
which then casts a challenging functional optimization problem.
Fortunately, it turns out that a simple closed form solution can be obtained by taking
$\mathcal{F}$ to be an RKHS $\mathcal{H}=\mathcal{H}_0\times\cdots \mathcal{H}_0$,
where $\H_0$ is a RKHS of scalar-valued functions, associated with a positive definite kernel $k(x,x')$.
In this case, \citet{liu2016kernelized} showed that the optimal solution  of \eqref{vgddecrease} is $\ff^* / ||\ff^* ||_\H$, where
 \begin{equation}
 \label{transf}
 \bd{\phi}^*(\cdot)
 =\mathbb{E}_{\bd{x}\sim{q}}[\nabla_{\bd{x}} \log p(\bd{x})k(\bd{x},\cdot)+\nabla_{\bd{x}} k(\bd{x},\cdot)].
 \end{equation}
In addition, the corresponding Stein discrepancy, known as kernelized Stein discrepancy (KSD) \citep{liu2016kernelized, chwialkowski2016kernel, gretton2009fast, oates2016control}, can be shown to have the following closed form
\begin{align}
\label{equ:sdefine}
\mathbb{D}(q ~||~ p) = ||\ff^* ||_\H =  \big(\mathbb{E}_{x,x'\sim q}[\kappa_p(\vx,\vx')]\big)^{1/2},
\end{align}
where $\kappa_p(x,x')$ is a positive definite kernel defined by
 \begin{align}
\kappa_p & (\bd{x},  \bd{x}') 
 = \bd{s}_p(\bd{x})^\top k(\bd{x},\bd{x}')\bd{s}_p(\bd{x}') + \bd{s}_p(\bd{x})^\top \nabla_{\bd{x}'}k(\bd{x},\bd{x}') \notag\\
& +\bd{s}_p(\bd{x}')^\top \nabla_{\bd{x}} k(\bd{x},\bd{x}')+\nabla_{\bd{x}}\cdot(\nabla_{\bd{x}'}k(\bd{x}, \bd{x}')). \label{ksdkernel}
\end{align}
where $\bd s_p(\vx) \overset{def}{=} \nabla \log p(\vx)$. 
We refer to \citet{liu2016kernelized} 
for the  derivation of \eqref{ksdkernel}, and further treatment of KSD in  
\citet{chwialkowski2016kernel, oates2016control, gorham2017measuring}.

\subsection{Stein Variational Gradient Descent}
In order to apply the derived optimal transform in the practical SVGD algorithm, we approximate the expectation $\E_{\vx\sim q}[\cdot]$ in \eqref{transf} using  the empirical averaging of the current particles,
that is, given particles $\{\vx_i^\ell\}_{i=1}^n$ at the $\ell$-th iteration, we construct the following velocity field:
\begin{align}\label{equ:phit}
\!\!\! \!  \ff_{\ell+1}(\cdot ) =\frac{1}{n} \sum_{j=1}^n [\nabla \log p(\bd{x}_j^{\ell})k(\bd{x}_j^{\ell},\cdot)+\nabla_{\bd{x}_j^{\ell}} k(\bd{x}_j^{\ell},\cdot)].
\end{align}
The SVGD update at the $\ell$-th iteration is then given by
%
\begin{align}\label{equ:svgdup2}
\begin{split}
& \vx_i^{\ell+1}  \gets \T_{\ell+1}(\vx_{i}^{\ell}), \\
& \T_{\ell+1}(\vx) =\vx + \epsilon \ff_{\ell+1}(\vx).
\end{split}
\end{align}
Here transform $\T_{\ell+1}$ is adaptively constructed based on the most recent particles $\{\vx_i^\ell\}_{i=1}^n$.
 Assume the initial particles $\{\vx_i^0\}_{i=1}^n$ are i.i.d. drawn from some distribution $q_0$, then the pushforward maps of $\T_\ell$ define a sequence of distributions that bridges between $q_0$ and $p$:
\begin{align}\label{equ:qt}
q_{\ell} =  (\T_{\ell} \circ \cdots \circ \T_{1})\sharp q_0, \quad \ell=1, \ldots, K,
\end{align}
where $q_\ell$ forms increasingly better approximation of the target $p$ as $\ell$ increases.
Because $\{\T_\ell\}$ are nonlinear transforms,
$q_\ell$ can represent highly complex distributions even when the original $q_0$ is simple.
In fact, one can view $q_\ell$ as
a deep residual network \citep{he2016deep} constructed layer-by-layer in a fast, nonparametric fashion.

However,
because the transform $\T_{\ell}$ depends on the previous particles $\{\vx_{i}^{\ell-1}\}_{i=1}^n$ as shown in \eqref{equ:phit},
the particles $\{\vx_i^{\ell}\}_{i=1}^n$, after the zero-th iteration, depend on each other in a complex fashion,
and do not, in fact, straightforwardly follow distribution $q_{\ell}$ in \eqref{equ:qt}.
Principled approaches for analyzing such interacting particle systems can be found in~\citet[e.g.,][]{braun1977vlasov, spohn2012large, del2013mean, sznitman1991topics}. The goal of this work, however, is to provide a simple method to ``decouple'' the SVGD dynamics, transforming it into a standard 
importance sampling method that is amendable to easier analysis and interpretability, and also applicable to more general inference tasks such as estimating partition function of unnormalized distribution where SVGD cannot be applied.



\section{DECOUPLING SVGD}
In this section, we introduce our main Stein variational importance sampling~(SteinIS) algorithm. Our idea is simple.
We initialize the particles $\{\vx_i^0\}_{i=1}^n$ by i.i.d. draws from an initial distribution $q_0$ and partition them into two sets,
including a set of \emph{leader particles} $\vx_A^\ell = \{\vx_i^\ell  \colon  i\in A\}$
and
 \emph{follower particles}
$\vx_B^\ell = \{\vx_i^\ell  \colon  i\in B\}$, with $B =  \{1,\ldots, n\} \setminus A$,
where the leader particles $\vx_A^\ell$ are responsible for constructing the transforms, using the standard SVGD update \eqref{equ:svgdup2},
while the follower particles $\vx_B^\ell$ simply follow the transform maps constructed by $\vx_A^\ell$ and do not contribute to the construction of the transforms. 
In this way, the follower particles $\vx_B^{\ell}$ are independent conditional on the leader particles $\vx_A^\ell$. 

Conceptually, we can think that we first construct all the maps $\T_\ell$ by evolving the leader particles $\vx_A^\ell$,
and then push the follower particles through $\T_\ell$ in order to draw exact, i.i.d. samples from $q_\ell$ in \eqref{equ:qt}.
Note that this is under the assumption the leader particles $\vx_A^\ell$
has been observed and fixed, which is necessary because the transform $\T_\ell$ and
distribution $q_\ell$ depend on $\vx_A^\ell$.
\begin{figure}[tbp]
   \centering
 \includegraphics[width=.32\textwidth]{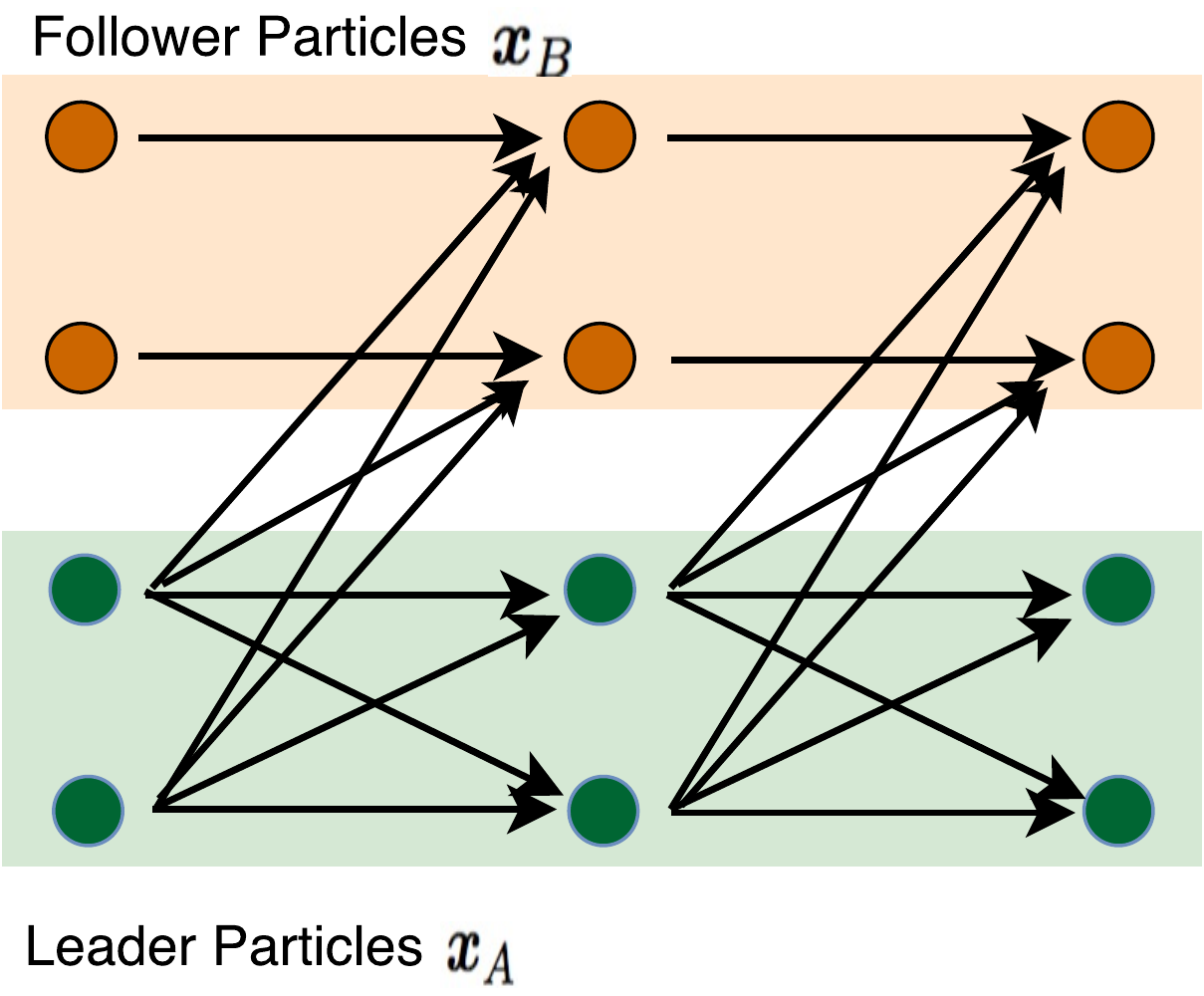} 
   \vspace{-.45\baselineskip}
   \caption{\small
   Our method uses a set of leader particles $\vx_A^\ell$ (green) to construct the transform map $\T_\ell$,
   which follower particles $\vx_B^\ell$ follows subsequently. The leader particles $\vx_A^\ell$ are interactive and dependent on each other.
   The follower particles $\vx_B^\ell$ can be viewed as i.i.d. draws from $q_\ell$, given fixed leader particles $\vx_A^\ell$.
   }
   \label{fig:example}
\end{figure}

In practice, however, we can simultaneously update both the leader and follower particles, by a simple modification of the original SVGD \eqref{equ:svgdup2} shown in Algorithm~\ref{VarIS:algo} (step 1-2),
where the only difference is that we restrict the empirical averaging in \eqref{equ:phit} to the set of 
the leader particles $\vx_A^\ell$. The relationship between the particles in set $A$ and $B$ can be more easily understood in Figure~\ref{fig:example}.
\begin{algorithm}[tb] %
\caption{Stein Variational Importance Sampling}  \label{VarIS:algo}
\begin{algorithmic}
\STATE {\bf Goal}: Obtain i.i.d. importance sample $\{\vx_i^K, ~ w_i^K\}$ for $p$.
\STATE {Initialize} $\vx_{A}^0$ and $\vx_B^0$ by i.i.d. draws from $q_0$.
\STATE {Calculate} $\{ q_0(\vx_i^0) \}, \forall i\in B.$
\FOR{iteration $\ell =0, \ldots, K-1$}
\STATE
1. Construct the map using the leader particles $\vx_A^\ell$
\begin{align}\notag
 \!\!\!\! \!\!\!\!
\ff_{\ell+1}(\cdot) =  \frac{1}{|A|}\sum_{j\in A} [\nabla \log p(\vv x_j^\ell) k(\vv x_j^\ell,  \cdot) + \nabla_{\vv x_j^\ell} k(\vv x_j^\ell, \cdot)].
\end{align}
2. Update both the leader and follower particles 
$$
\vv x_{i}^{\ell+1} \gets \vv x_i^\ell + \epsilon \ff_{\ell+1}(\vv x_i^\ell), ~~~~ \forall i \in A\cup B.
$$
3. Update the density values (for $i\in B$) by
\begin{equation*}
q_{\ell+1}(\vx_{i}^{\ell+1}) =  q_{\ell}(\vx_{i}^{\ell}) \cdot  | \mathrm{det}(I ~+~ \epsilon \nabla_{\bd{x}} \bd{\ff}_{\ell+1}(\bd{x}^{\ell}_i))|^{-1}
\end{equation*}
\ENDFOR
\STATE {Calcuate} $w_i^{K} = p(\vv x_i^{K}) / q_K(\vv x_i^{K}), \forall i\in B.$
\STATE {\bf Outputs}: i.i.d. importance sample $\{\vx_i^K, ~ w_i^K\}$ for $i\in B.$
\end{algorithmic}
\end{algorithm}

\paragraph{Calculating the Importance Weights}
Because $q_\ell$ is still different from $p$ when we only apply finite number of iterations $\ell$, which introduces deterministic biases if we directly use $\vx_B^\ell$ to approximate $p$.
We address this problem by further turning the algorithm into an importance sampling algorithm with importance proposal $q_\ell$. Specifically, we calculate the importance weights of the particles $
\{\vx_i^\ell\}$:
\begin{align}\label{equ:wt}
w_i^\ell= \frac{\bar p(\vx_i^{\ell})}{q_\ell(\vx_i^\ell)},
\end{align}
where $\bar p$ is the unnormalized density of $p$, that is, $p(\vx) = \bar p(\vx)/Z$ as in \eqref{barp}. 
In addition, the importance weights in \eqref{equ:wt} can be calculated based on the following formula:
\begin{equation}
\label{density}
q_\ell(\bd{x}^\ell)=q_0(\bd{x}^0)\prod_{\jmath=1}^\ell |\mathrm{det}(\nabla_{\bd{x}} \bd{T}_\jmath(\bd{x}^{\jmath-1}))|^{-1},
\end{equation}
where $\bd{T}_\ell$ is defined in~\eqref{equ:svgdup2} and we assume that
 the step size $\epsilon$ is small enough so that each $\T_\ell$ is an one-to-one map.
 As shown in Algorithm~\ref{VarIS:algo} (step 3), 
 \eqref{density} can be calculated recursively as we update the particles. 

With the importance weights calculated, we turn SVGD into a standard importance sampling algorithm. 
For example, we can now estimate expectations of form $\E_p f$ by 
$$
 \hat \E_{p}[f] = \frac{\sum_{i\in B} w_i^\ell f(\vx_i^\ell)}{\sum_{i\in B} w_i^\ell}, 
$$
which provides a consistent estimator of $\E_{p} f$ when we use finite number $\ell$ of transformations.  
Here we use the self normalized weights because
$\bar p(\vx)$ is unnormalized.
Further, the sum of the unnormalized weights provides an unbiased estimation for the normalization constant $Z$:
$$
\hat Z = \frac{1}{|B|} \sum_{i\in B} w_i^\ell,
$$
which satisfies the unbiasedness property $\E[\hat Z] = Z$. Note that the original SVGD does not provide a method for estimating normalization constants,
although, as a side result of this work, Section 4 will discuss another method for estimating $Z$ that is more directly motivated by SVGD.



We now analyze the time complexity of our algorithm. 
Let $\alpha(d)$ be the cost of computing $\bd{s}_p(\bd{x})$ and 
$\beta(d)$ be the cost of evaluating kernel $k(\vx, \vx')$ and its gradient $\nabla k(\vx, \vx')$. Typically, both $\alpha(d)$ and $\beta(d)$ grow linearly with the dimension $d. $ In most cases, $\alpha(d)$ is much larger than $\beta(d)$. 
The complexity of the original SVGD with $|A|$ particles is $O(|A|\alpha(d)+|A|^2\beta(d))$, 
and the complexity of Algorithm~\ref{VarIS:algo} is $O(|A|\alpha(d)+|A|^2\beta(d)+|B||A|\beta(d)+|B| d^3 ),$   
where the $O(|B|d^3)$ complexity comes from calculating the determinant of the Jacobian matrix, 
which is expensive when dimension $d$ is high, but is the cost to pay for having a consistent importance sampling estimator in finite iterations
and for being able to estimate the normalization constant $Z$.
Also, by calculating the effective sample size based on the importance weights, 
we can assess the accuracy of the estimator, and early stop the algorithm when a confidence threshold is reached. 

One way to speed up our algorithm in empirical experiments is to parallelize the computation of Jacobian matrices for all follower particles in GPU. It is possible, however, to develop efficient approximation for the determinants by leveraging the special structure of the Jacobean matrix; note that 
\begin{align}
&\nabla_{\bd{y}} \bd{T}(\bd{y}) = I  + \epsilon A,  \notag \\
&A = \frac{1}{n}\sum_{j=1}^n [\nabla_{\vv x} \log p(\vv x_j)^\top \nabla_{\vv y} k(\vv x_j, \vv y) +
\nabla_{\vv x} \nabla_{\vv y} k(\vv x_j, \vv y)]. \notag
 \end{align}
 Therefore, $\nabla_{\bd{y}} \bd{T}(\bd{y})$ 
 is close to the identity matrix $I$ when the step size is small.
This allows us to use Taylor expansion for approximation:
\begin{pro}
\label{detapprox}
Assume $\epsilon < 1/\rho(A)$, where $\rho(A)$ is the spectral radius of $A$, that is,
 $\rho(A) =\max_j |\lambda_j(A)|$ and $\{\lambda_j\}$ are the eigenvalues of $A$. We have
\begin{equation}
\label{Jacobapprox}
\mathrm{det}(I +\epsilon A) =\prod_{k=1}^d (1+\epsilon a_{kk})+ O(\epsilon^2),
\end{equation}
where $\{a_{kk}\}$ are the diagonal elements of $A$.
\end{pro}

\begin{proof}
Use the Taylor expansion of $\mathrm{det}(I +\epsilon A)$. 
\end{proof}
Therefore, one can approximate the determinant with approximation error $O(\epsilon^2)$ using linear time $O(d)$ w.r.t. the dimension.
Often the step size is decreasing with iterations,
and a way to trade-off the accuracy with computational cost is to 
use the exact calculation in the beginning when the step size is large,
and switch to the approximation when the step size is small.
\subsection{Monotone Decreasing of KL divergence} 
One nice property of algorithm~\ref{VarIS:algo} is that the KL divergence between the iterative distribution $q_\ell$ and $p$ is monotonically decreasing. This property can be more easily understood by considering our iterative system in continuous evolution time as shown in \citet{liu2017stein}. 
Take the step size $\epsilon$ of the transformation defined in \eqref{update} to be infinitesimal, 
and define the continuos time $t = \epsilon \ell$. Then the evolution equation of random variable $\vx^t$ is governed by the following nonlinear partial differential equation~(PDE), 
\begin{equation}
\label{part}
\frac{d\bd{x}^t}{dt}=\mathbb{E}_{\bd{x}\sim{q_t}}[\bd{s}_p(\bd{x})k(\bd{x},\bd{x}^t)+\nabla_{\bd{x}} k(\bd{x},\bd{x}^t)],
\end{equation}
where $t$ is the current evolution time and $q_t$ is the density function of $\vx^t.$ The current evolution time $t= \epsilon \ell$ when $\epsilon$ is small and $\ell$ is the current iteration. We have the following proposition (see also \citet{liu2017stein}): 
\begin{pro}
\label{pro2}
Suppose random variable $\bd{x}^t$ is governed by PDE \eqref{part}, then its density $q_t$ is characterized by
\begin{equation}
\label{diffode}
\frac{\partial q_t}{\partial t}=-\mathrm{div}(q_t\mathbb{E}_{\bd{x}\sim{q_t}}[\bd{s}_p(\bd{x}) k(\bd{x},\bd{x}^t)+\nabla_{\bd{x}} k(\bd{x},\bd{x}^t)]),
\end{equation}
where $\mathrm{div}(\bd{f})=\trace(\nabla \vv f) = \sum_{i=0}^d \partial f_i(\bd{x})/\partial x_i$, and  $\bd{f}=[f_1,\ldots, f_d]^\top.$ 
\end{pro}
The proof of proposition~\ref{pro2} is similar to the proofs of proposition 1.1 in~\citet{jourdain1998propagation}. 
Proposition~\ref{pro2} characterizes the evolution of the density function $q_t(\bd{x}^t)$ when the random variable $\bd{x}^t$ is evolved by ~\eqref{part}. The continuous system captured by~\eqref{part} and ~\eqref{diffode} is a type of Vlasov process which has wide applications in physics, biology and many other areas~\citep[e.g.,][]{braun1977vlasov}.
As a consequence of proposition~\ref{pro2}, one can show the following nice property:
\begin{equation}
\label{klksd}
\frac{d\mathrm{KL}(q_t\mid\mid p)}{dt}=-\mathbb{D}(q_t ~||~ p)^2<0,
\end{equation}
which is proved by theorem 4.4 in \citet{liu2017stein}. 
Equation ~\eqref{klksd} indicates that the KL divergence between the iterative distribution $q_t$ and $p$ is monotonically decreasing with a rate of $\mathbb{D}(q_t~||~ p)^2$. 

\section{A PATH INTEGRATION METHOD}
\begin{algorithm}[tb]
\caption{SVGD with Path Integration for estimating $\KL(q_0 ~||~p)$ and $\log Z$}
\label{ksdalgo}
\begin{algorithmic}[1]
\STATE {\bfseries Input:} Target distribution $p(x) = \bar p(x)/Z$; an initial distribution $q_0$.
\STATE {\bfseries Goal:} Estimating $\KL(q_0\mid\mid p)$ and the normalization constant $\log Z.$
\STATE{ Initialize $\hat K =0.$ Initialize particles $\{\vx_i^0\}_{i=1}^n\sim q_0(\bd{x}).$}
\STATE{Compute $\hat{\mathbb{E}}_{q_0}[\log(q_0(\vx)/\overline{p}(\vx))]$ via sampling from $q_0.$}
\WHILE{iteration $\ell$}
\STATE 
$$
\hat K \gets \hat K + \epsilon  \hat{\mathbb{D}}({q}_\ell ~||~ p)^2,
$$
$$\vx_i^{\ell+1} \gets \vx_i^\ell + \bd{\phi}_{\ell +1}(\vx_i^\ell), $$
where $\hat{\mathbb{D}}({q}_\ell ~||~ p)$ is defined in \eqref{equ:hats}.
\ENDWHILE
\STATE {Estimate $\KL(q_0 ~||~  p)$ by $\hat K$ and $\log Z$ by $\hat{\mathbb{D}} - \hat{\mathbb{E}}_{q_0}[\log(q_0(\vx)/\overline{p}(\vx))].$}
\end{algorithmic}
\end{algorithm}

We mentioned that the original SVGD
does not have the ability to estimate the partition function. 
Section 3 addressed this problem by 
turning SVGD into a standard importance sampling algorithm in Section 3.
Here we introduce another method
for estimating KL divergence and normalization constants that is more directly motivated by the original SVGD,
by leveraging the
fact that the Stein discrepancy is a type of gradient of KL divergence.
This method does not need to estimate the importance weights but has to run SVGD to converge to diminish the Stein discrepancy between intermediate distribution $q_\ell$ and $p$. In addition, this method does not perform as well as Algorithm 1 as we find empirically.
Nevertheless, we find this idea is conceptually interesting and useful to discuss it.

Recalling Equation {\eqref{vgddecrease}} in Section 2.1, we know that
if we perform transform $\T(\vx) = \vx +  \epsilon \ff^*(\vx)$ with $\ff^*$ defined in \eqref{transf},
the corresponding decrease of KL divergence would be
\begin{align}
\begin{split}
 \KL(q ~||~ p)  - \KL(\T\sharp q  ~||~ p)
&  \approx \epsilon \cdot || \ff^* ||_\H \cdot \S(q ~|| ~ p)  \\
&  \approx \epsilon \cdot \S(q ~|| ~ p)^2,
\end{split}
\label{def:kldecrease}
\end{align}
where we used the fact that $\S(q~||~p) = ||\ff^*||_\H$, shown in \eqref{equ:sdefine}.
Applying this recursively on $q_\ell$ in \eqref{def:kldecrease}, we get
$$
 \KL(q_0 ~||~ p) - \KL(q_{\ell+1} ~||~ p)  \approx \sum_{\jmath = 0}^\ell
\epsilon \cdot \S(q_{\jmath} ~|| ~ p)^2.
$$
Assuming $\KL(q_{\ell} ~||~ p) \to 0$ when $\ell \to \infty$,
we get
\begin{equation}
\label{klestimate}
\KL(q_0 ~||~ p)  \approx \sum_{\ell = 0}^\infty
\epsilon \cdot \S(q_{\ell} ~|| ~ p)^2.
\end{equation}

By \eqref{equ:sdefine}, the square of the KSD can be empirically estimated via V-statistics, which is given as
\begin{align}\label{equ:hats}
\hat \S(q_{\ell} ~|| ~ p)^2 = \frac{1}{n^2}\sum_{i,j=1}^n \kappa(\vx_i^\ell, \vx_j^\ell).
\end{align}
Overall, equation~\eqref{klestimate} and \eqref{equ:hats} give an estimator of the KL divergence between $q_0$ and $p = \bar p(\bd{x})/Z.$
This can be transformed into an estimator of the log normalization constant $\log Z$ of $p$, by noting that
\begin{equation}
\log Z=\KL(q_0\mid\mid p)-\mathbb{E}_{q_0}[\log(q_0(\vx)/\overline{p}(\vx))],
\end{equation}
where the second term can be estimated by drawing a lot of samples to diminish its variance since the samples from $q_0$ is easy to draw. 
The whole procedure is summarized in Algorithm~\ref{ksdalgo}.

\section{EMPIRICAL EXPERIMENTS}
\begin{figure}[t]
\centering
\begin{tabular}{cc}
\includegraphics[height=0.18\textwidth]{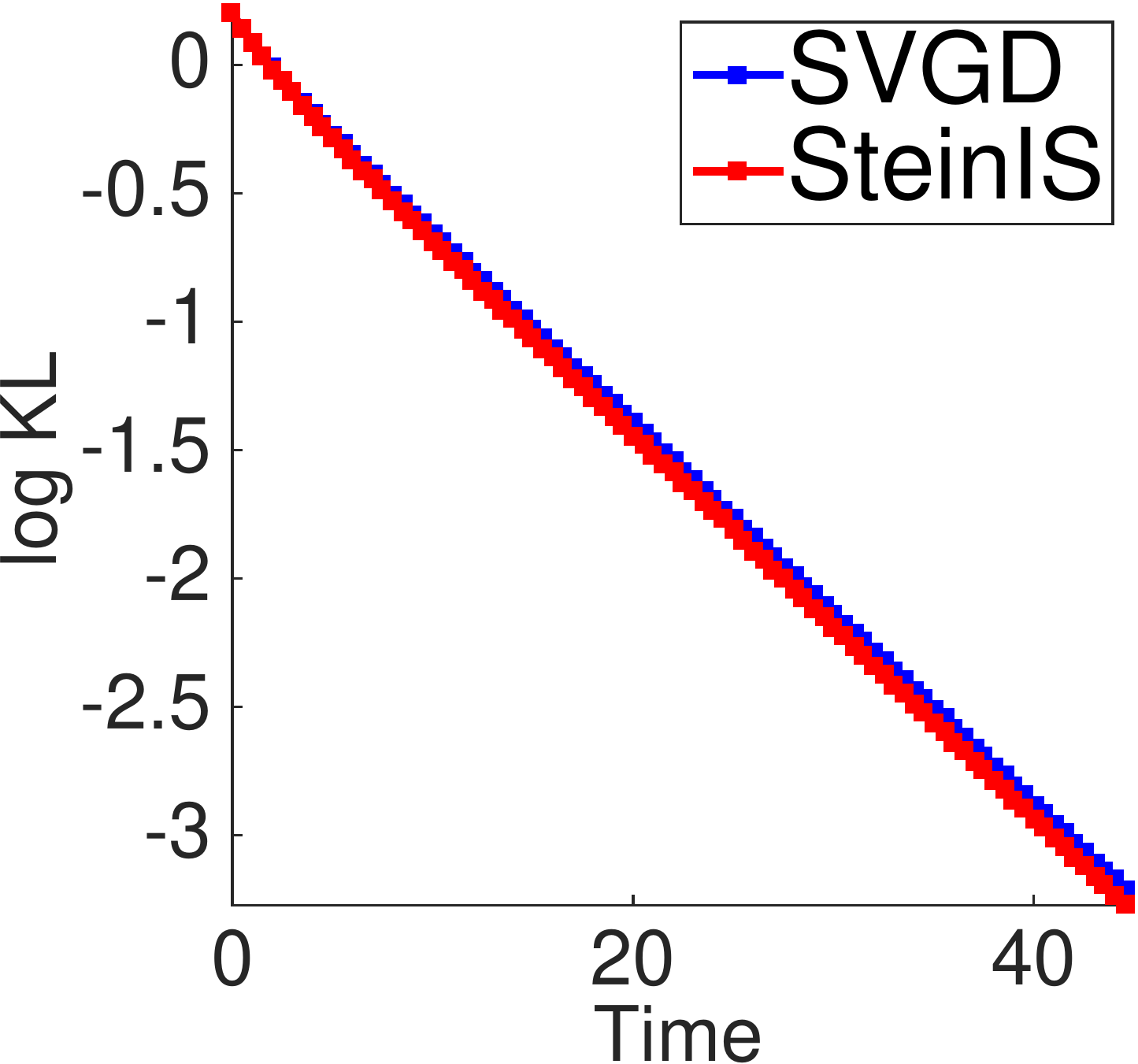} &
\includegraphics[height=0.18\textwidth]{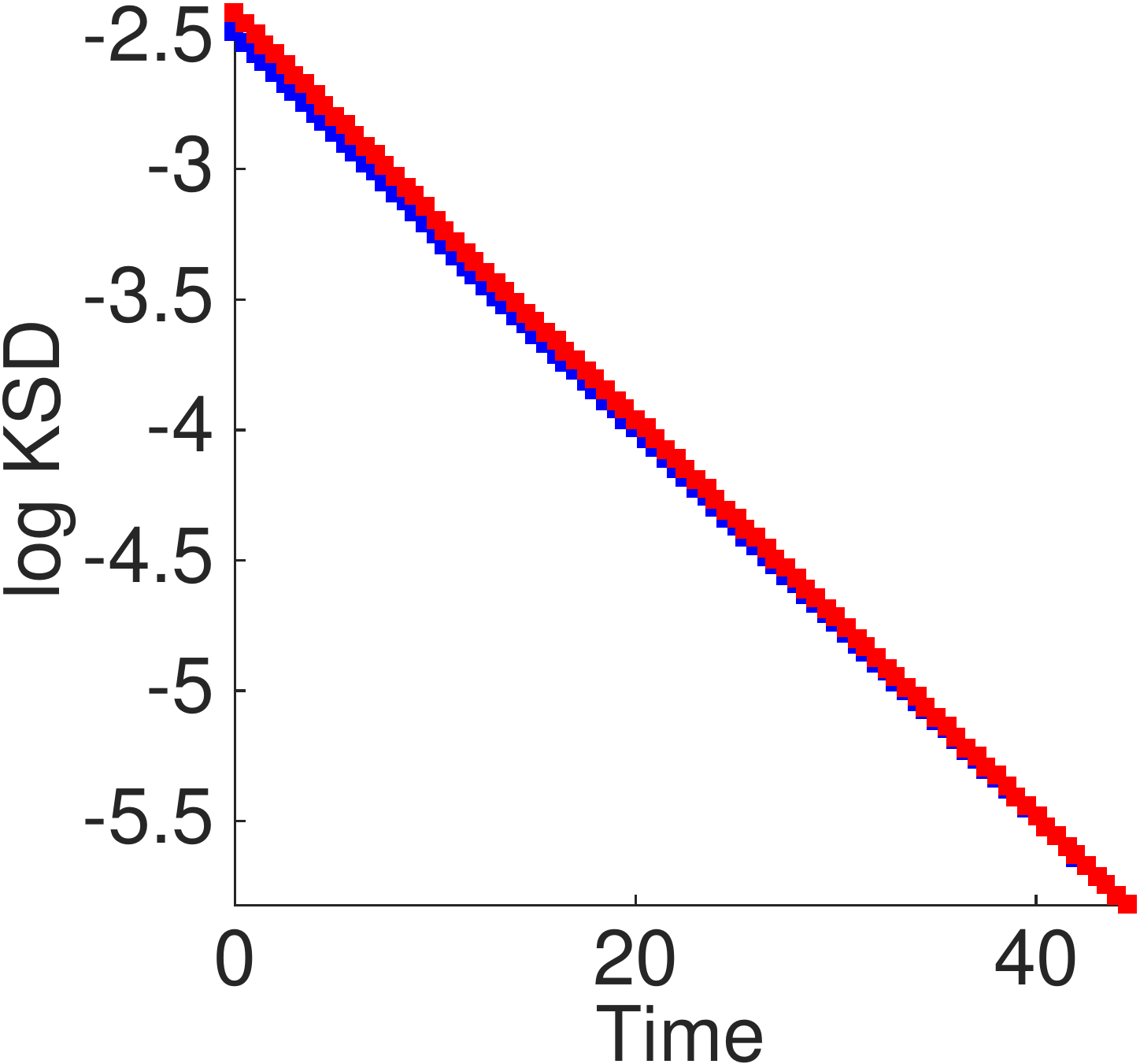} \\
{\small (a)  KL} &
{\small (b)  KSD}
\end{tabular}
\caption{GMM with 10 mixture components. $d=1.$ In SVGD, 500 particles are evolved. In SteinIS, $|A|=200$ and $|B|=500$. For SVGD and SteinIS, all particles are drawn from the same Gaussian distribution $q_0(\bd{x}).$}
\label{fig:KL}
\end{figure}

\begin{figure*}[t]
\centering
\begin{tabular}{cccc}
\includegraphics[height=0.19\textwidth]{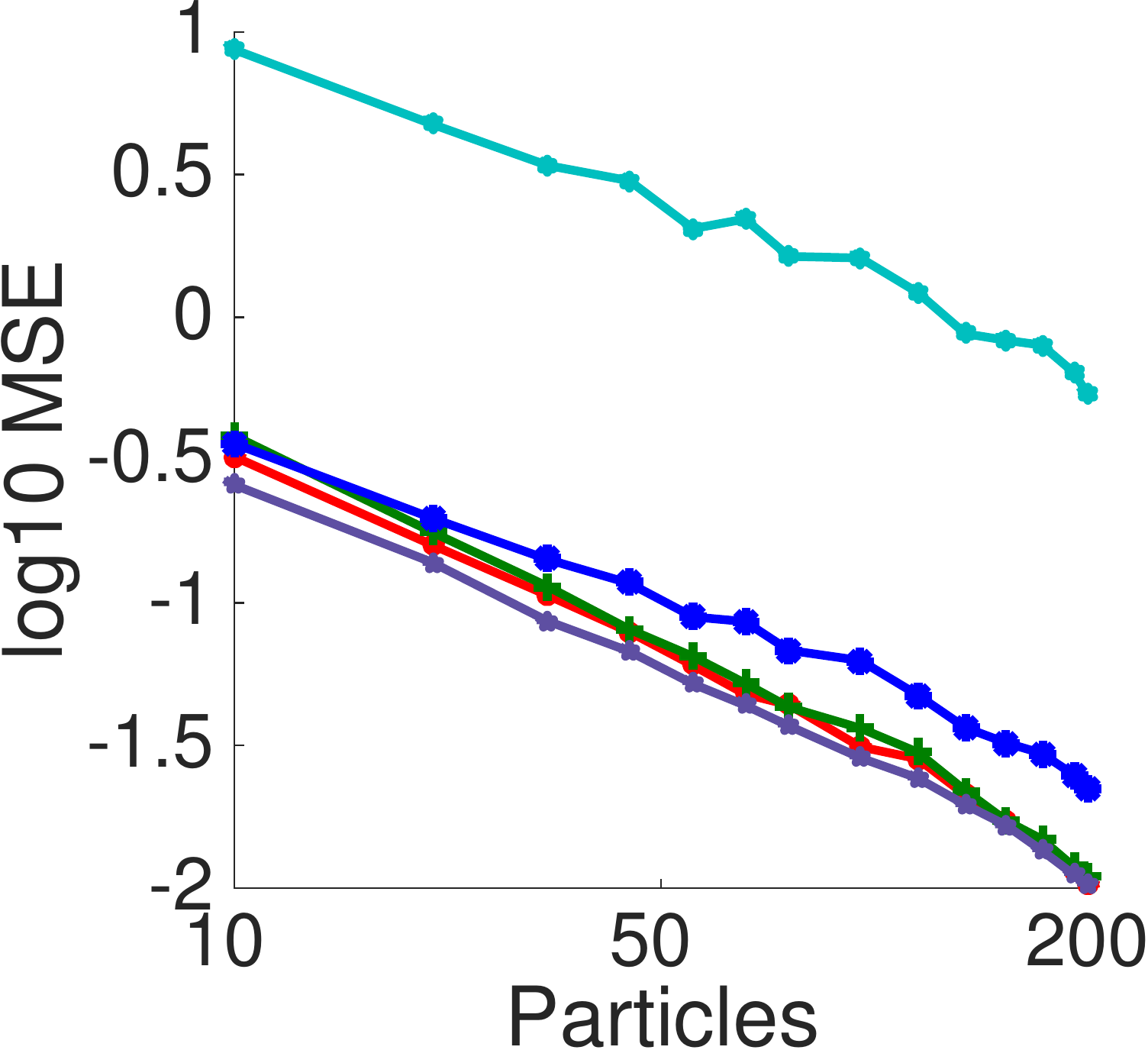} &
\includegraphics[height=0.19\textwidth]{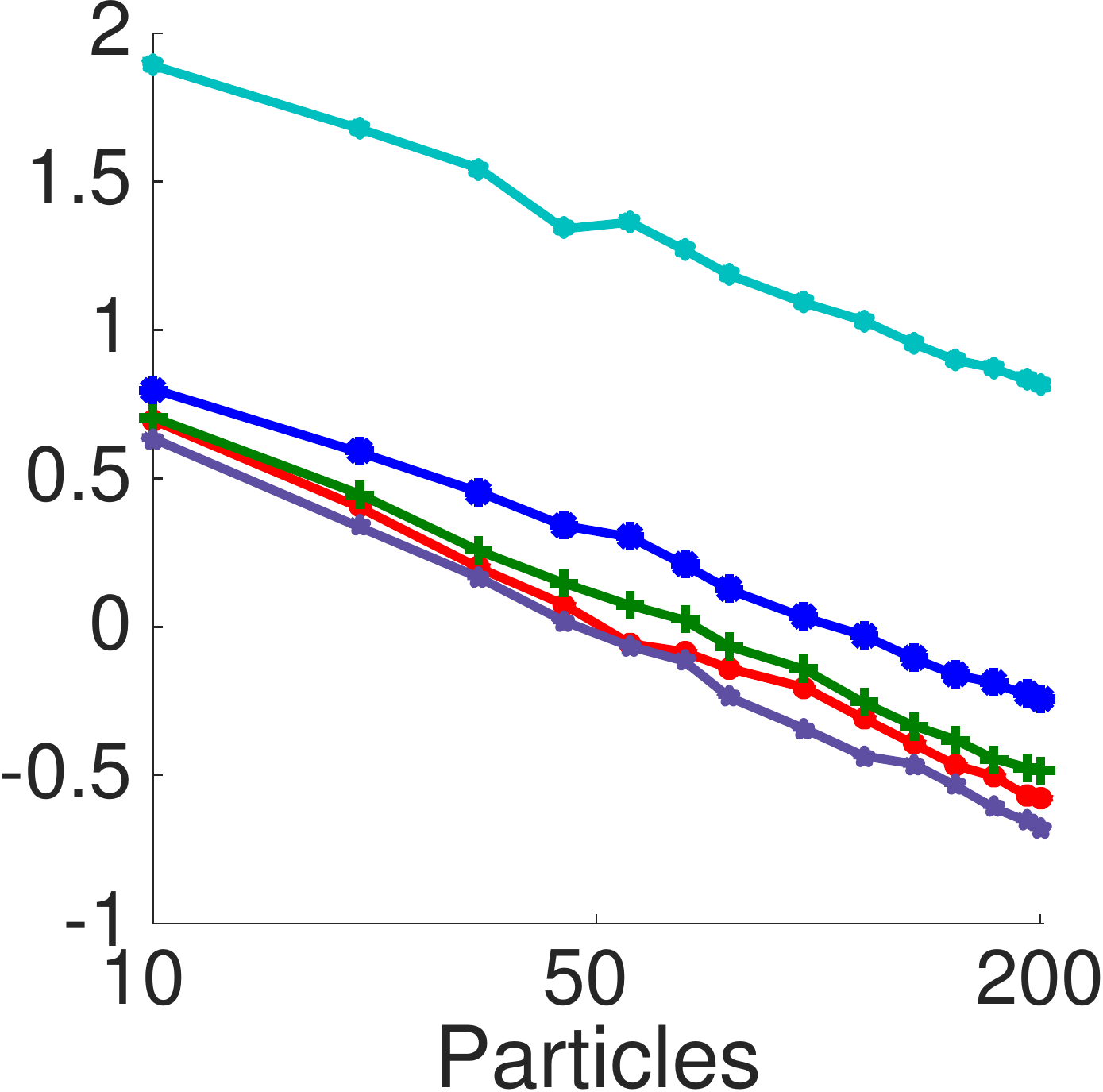} &
\includegraphics[height=0.19\textwidth]{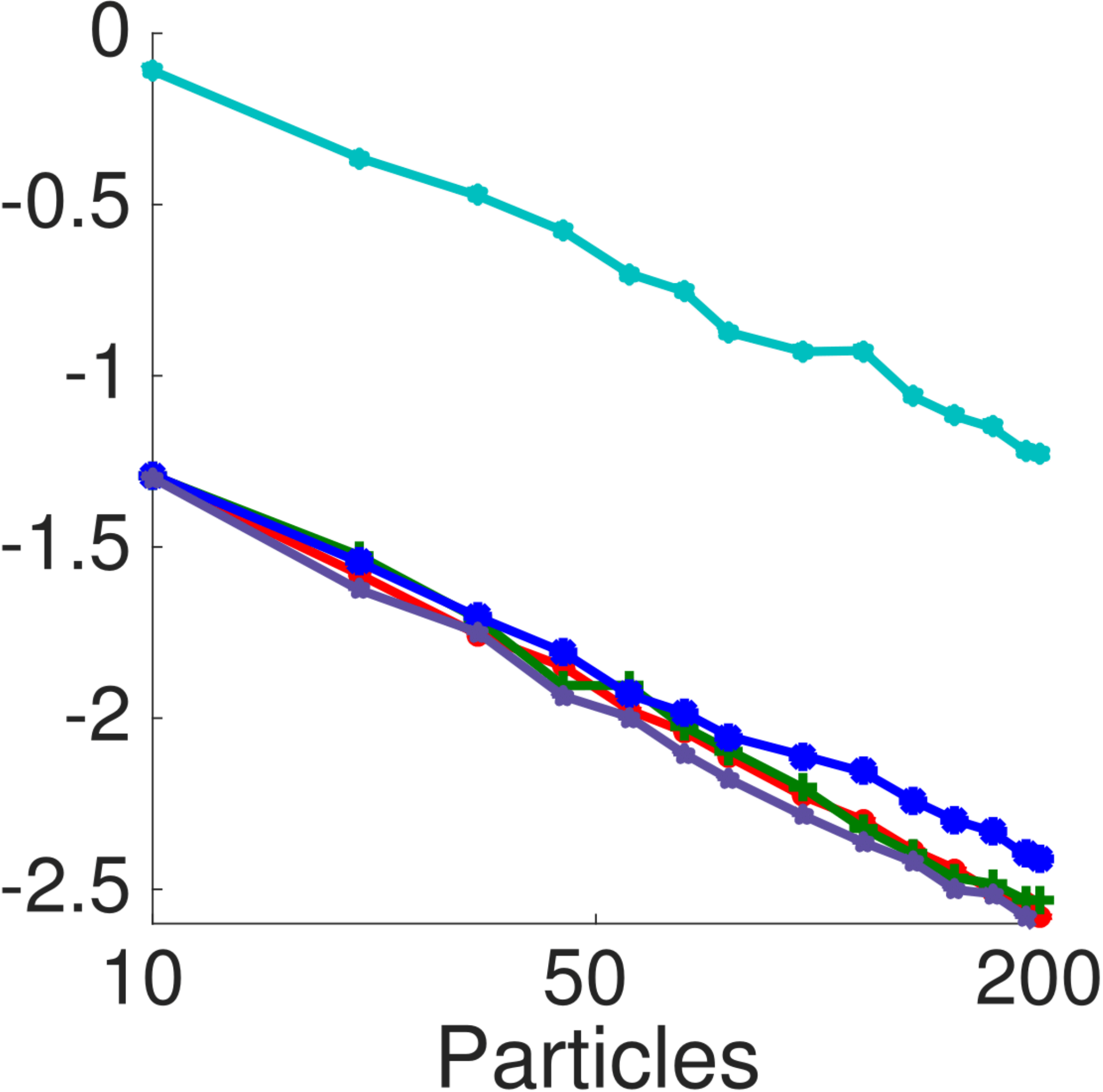} &
\includegraphics[height=0.19\textwidth]{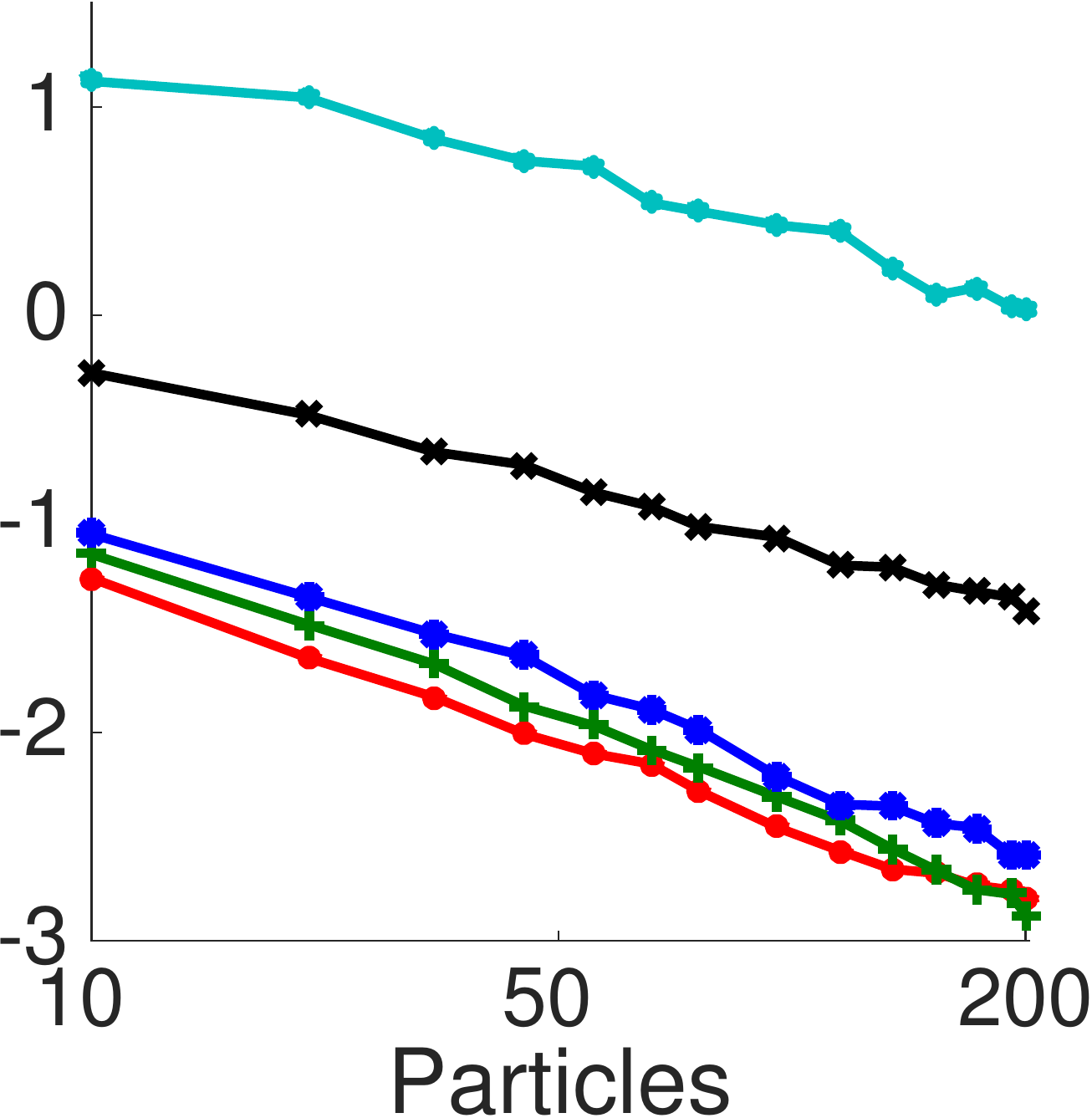}
\raisebox{2em}{ \includegraphics[height=0.1\textwidth]{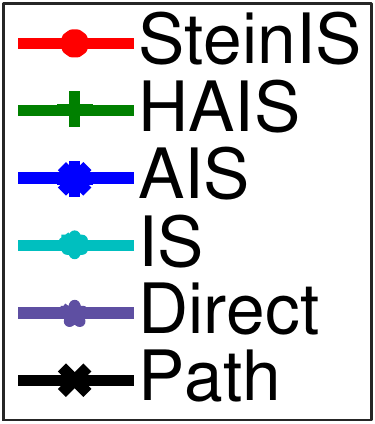}}\\
{\small (a)  $\mathbb{E}[x]$} &
{\small (b)  $\mathbb{E}[x^2]$ } &
{\small (c) $\mathbb{E}[\cos(w x + b)]$} &
{\small (d) Partition Function}\\
\end{tabular}
\caption{2D GMM with 10 randomly generated mixture components. (a)-(c) shows mean square error(MSE) for estimating $\mathbb{E}_p[h(x)],$ where $h(\bd{x})=x_j,~x_j^2,~\cos(wx_j+b)$ with $w\sim \normal(0,1)$ and $b\in \mathrm{Uniform}([0,1])$ for $j=1,2$, and the normalization constant (which is $1$ in this case). 
We used 800 transitions in SteinIS, HAIS and AIS, and take $L=1$ in HAIS. We fixed the size of the leader particles $|A|$ to be $100$ and vary the size of follower particles $|B|$ in SteinIS. The initial proposal $q_0$ is the standard Gaussian. "Direct" means that samples are directly drawn from $p$ and is not applicable in (d). "IS" means we directly draw samples from $q_0$ and apply standard importance sampling. "Path" denotes  path integration method in Algorithm~\ref{ksdalgo} and is only applicable to estimate the partition function in (d). The MSE is averaged on each coordinate over 500 independent experiments for SteinIS, HAIS, AIS and Direct, and over 2000 independent experiments for IS. SVGD has similar results (not shown for clarity) as our SteinIS on (a), (b), (c), but can not be applied to estimate the partition function in task (d). 
The logarithm base is 10.
}
\label{fig:varyparticle}
\end{figure*}

We study the empirical performance of our proposed algorithms on both simulated and real world datasets. 
We start with toy examples to numerically investigate some theoretical properties of our algorithms, 
and compare it with traditional adaptive IS on non-Gaussian, multi-modal distributions. 
We also employ our algorithm to estimate the partition function of Gaussian-Bernoulli Restricted Boltzmann Machine(RBM), a graphical model widely used in deep learning~\citep{welling2004exponential, hinton2006reducing}, and to evaluate the log likelihood of decoder models in variational autoencoder~\citep{kingma2013auto}.

We summarize some hyperparameters used in our experiments. We use RBF kernel $k(\bd{x}, \bd{x}')=\exp(-\|\bd{x}-\bd{x}'\|^2/h),$ where $h$ is the bandwidth. In most experiments, we let $h {=} \mathrm{med^2}/(2\log(|A|+1))$, where $\mathrm{med}$ is the median of the pairwise distance of the current leader particles $\vx_A^\ell$, 
and $|A|$ is the number of leader particles. 
 The step sizes in our algorithms are chosen to be $\epsilon =\alpha/(1+\ell)^\beta,$ where $\alpha$ and $\beta$ are hyperparameters chosen from a validation set to achieve best performance. When $\epsilon\le 0.1$, we use first-order approximation to calculate the determinants of Jacobian matrices as illustrated in proposition~\ref{detapprox}.

In what follows, 
we use ``{AIS}'' to refer to the annealing importance sampling with Langevin dynamics as its Markov transitions, and use ``{HAIS}'' to denote the annealing importance sampling whose Markov transition is Hamiltonian Monte Carlo (HMC). 
We use "transitions" to denote the number of intermediate distributions constructed in the paths of both SteinIS and AIS. 
A transition of HAIS may include $L$ leapfrog steps, as implemented by ~\citet{wu2016quantitative}. 

 \subsection{Gaussian Mixtures Models}
 We start with testing our methods on simple 2 dimensional Gaussian mixture models (GMM) with 10 randomly generated mixture components. 
First, we numerically investigate the convergence of KL divergence between the particle distribution $q_t$ (in continuous time) and $p.$ Sufficient particles are drawn and infinitesimal step $\epsilon$ is taken to closely simulate the continuous time system, as defined by \eqref{part}, \eqref{diffode} and \eqref{klksd}. Figrue~\ref{fig:KL}(a)-(b) show that the KL divergence $\KL(q_t, p)$, 
 as well as the squared Stein discrepancy $\mathbb{D}(q_t, p)^2$, seem to decay exponentially in both SteinIS and the original SVGD. 
 This suggests that the quality of our importance proposal $q_t$ improves quickly as we apply sufficient transformations. However, it is still an open question to establish the exponential decay theoretically; 
 see \citet{liu2017stein} for a related discussion. 

We also empirically verify the convergence property of our SteinIS 
as the follower particle size $|B|$ increases (as the leader particle size $|A|$ is fixed). We apply SteinIS to estimate $\mathbb{E}_p[h(x)],$ where $h(\bd{x})=x_j,~x_j^2 ~\textit{or}~\cos(wx_j+b)$ with $w \sim \normal(0,1)$ and $b \sim \mathrm{Uniform}([0,1])$ for $j=1,2$, and the partition function (which is trivially $1$ in this case). From Figure~\ref{fig:varyparticle}, we can see that the mean square error(MSE) of our algorithms follow the typical convergence rate of IS, which is $O(1/\sqrt{|B|}),$ where $|B|$ is the number of samples for performing IS. Figure~\ref{fig:varyparticle} indicates that SteinIS can achieve almost the same performance as the exact Monte Carlo~(which directly draws samples from the target $p$), 
indicating the proposal $q_\ell$ closely matches the target $p$. 

\begin{figure*}[t]
\centering
\scalebox{.99}{
\setlength{\tabcolsep}{0em}
\begin{tabular}{cccc}
\hspace{-.5cm} \includegraphics[height=0.22\textwidth]{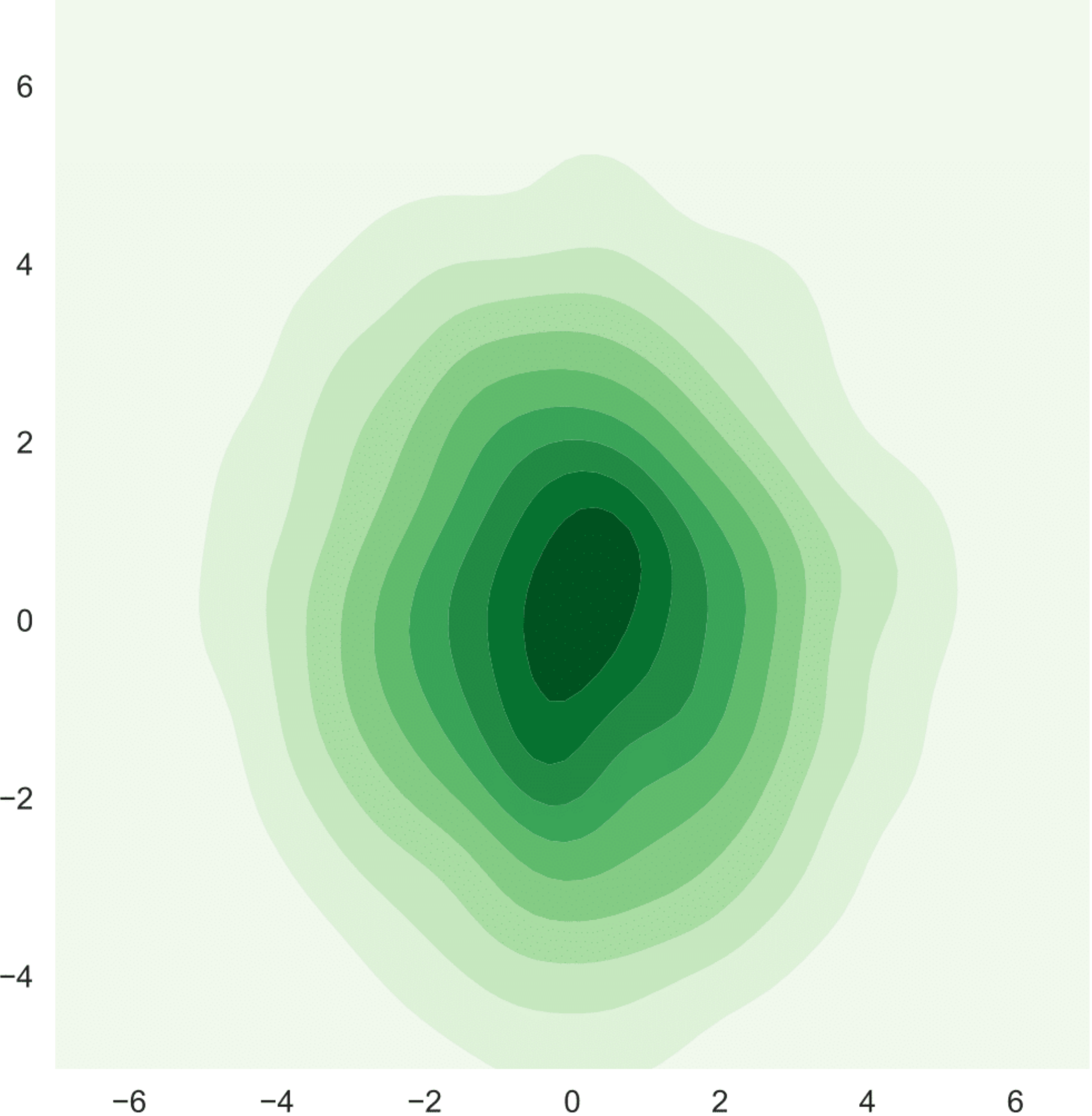}&
\includegraphics[height=0.22\textwidth]{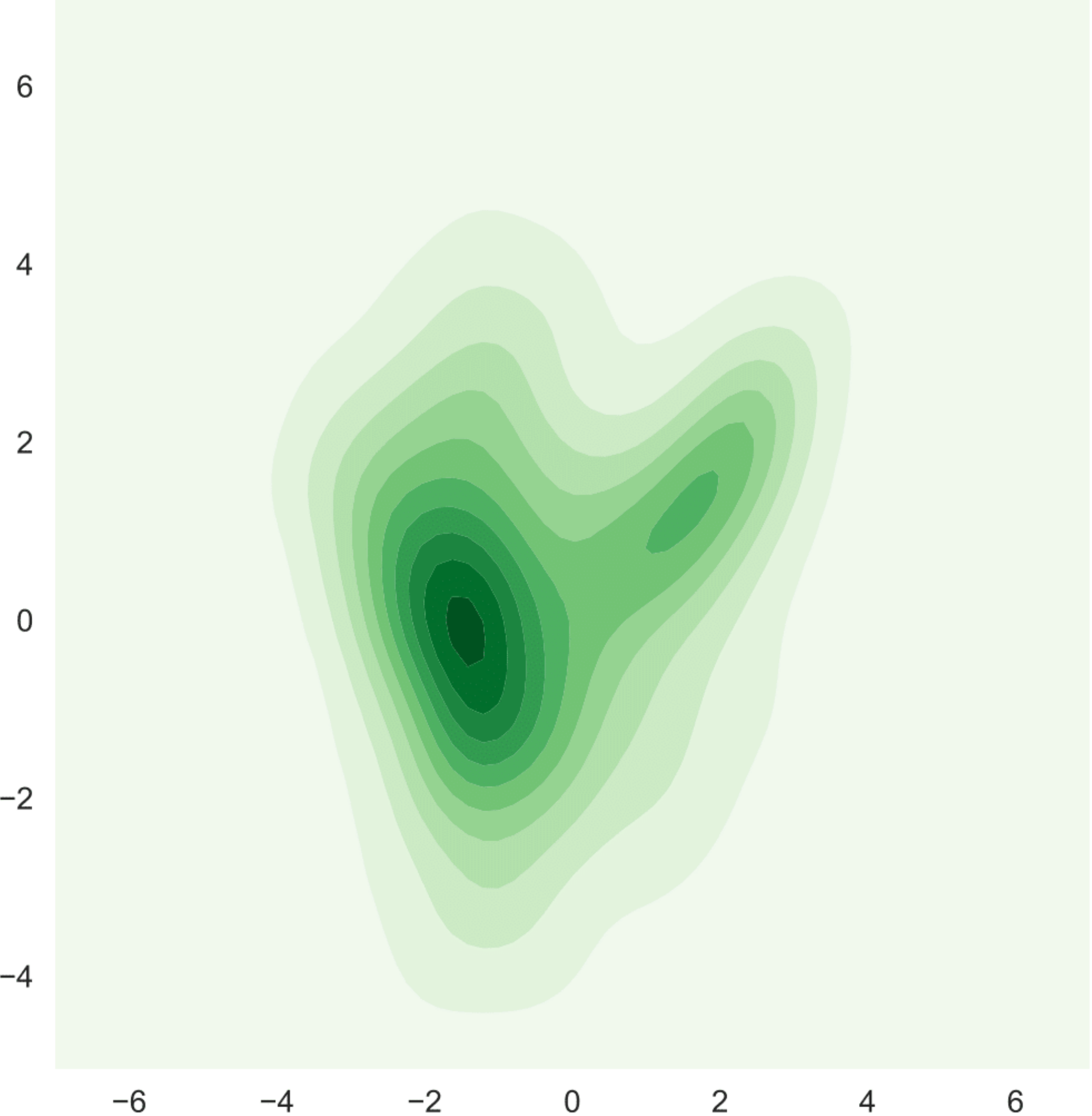} &
\includegraphics[height=0.22\textwidth]{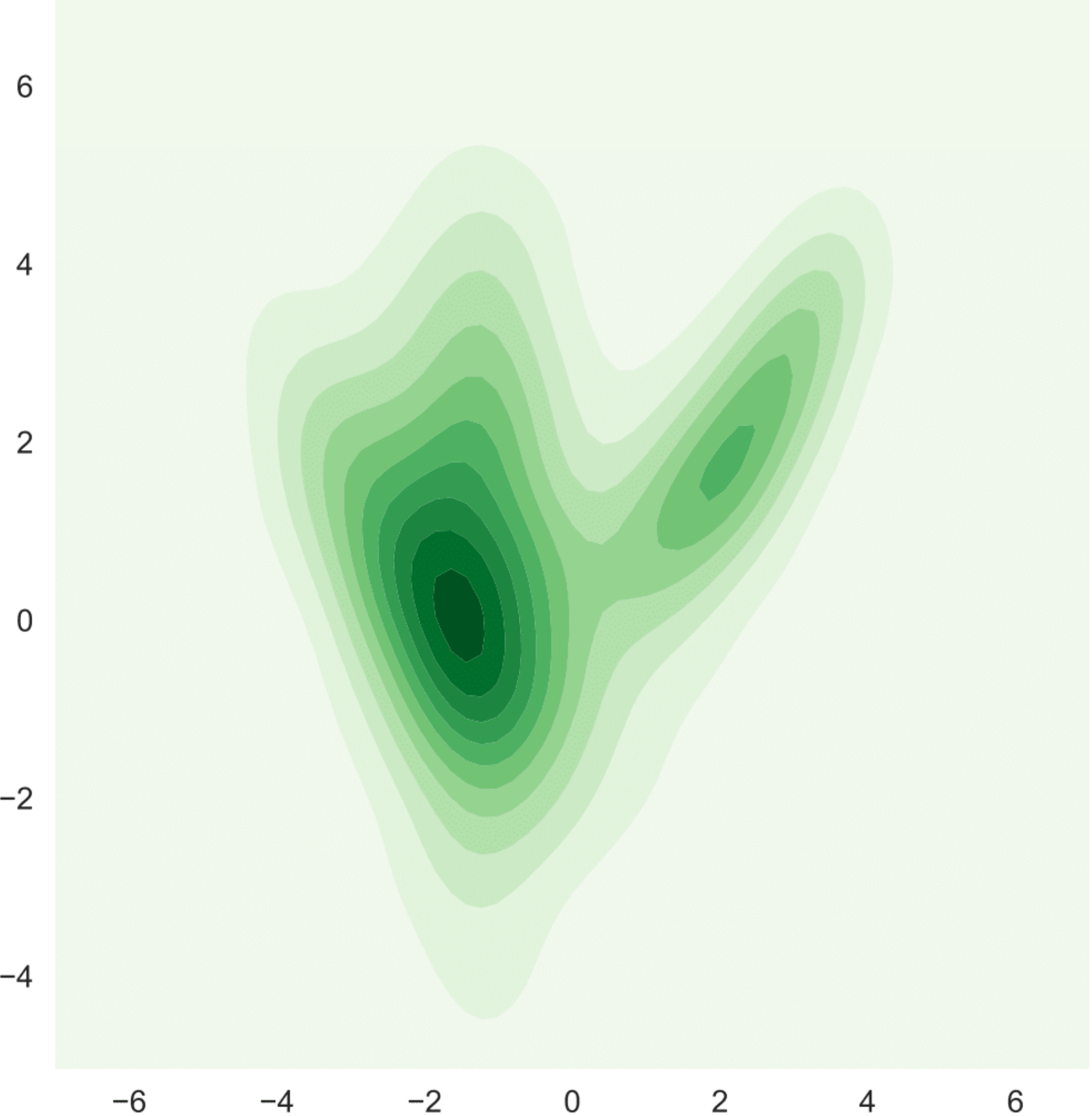} &
\includegraphics[height=0.22\textwidth]{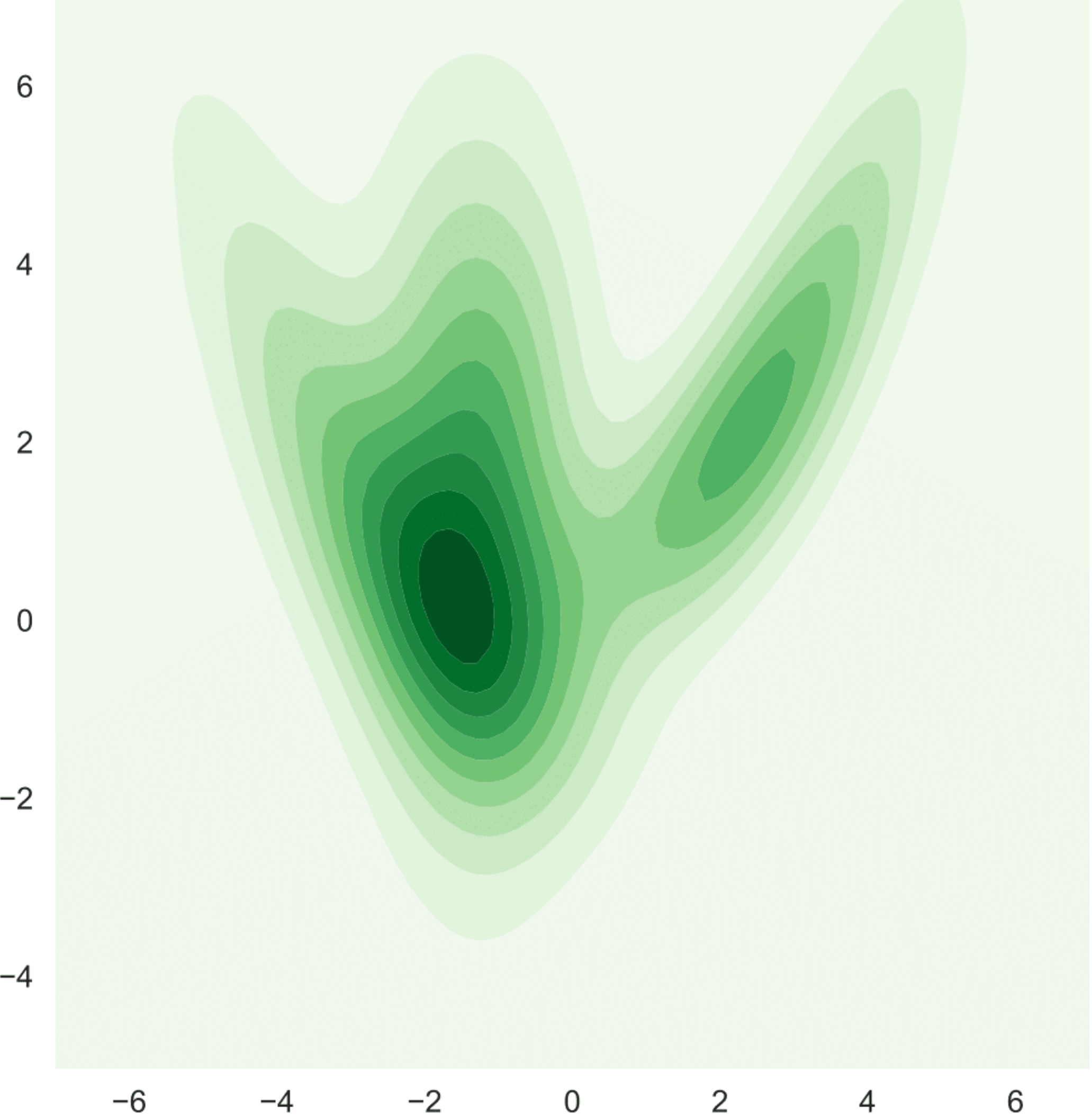} \\
{\small (a) SteinIS, $\ell=0$} & {\small (b) SteinIS,  $\ell=50$ } & {\small (c)  SteinIS, $\ell=200$} & {\small (d) SteinIS,  $\ell=2000$ } \\
\hspace{-.5cm} \includegraphics[height=0.21\textwidth]{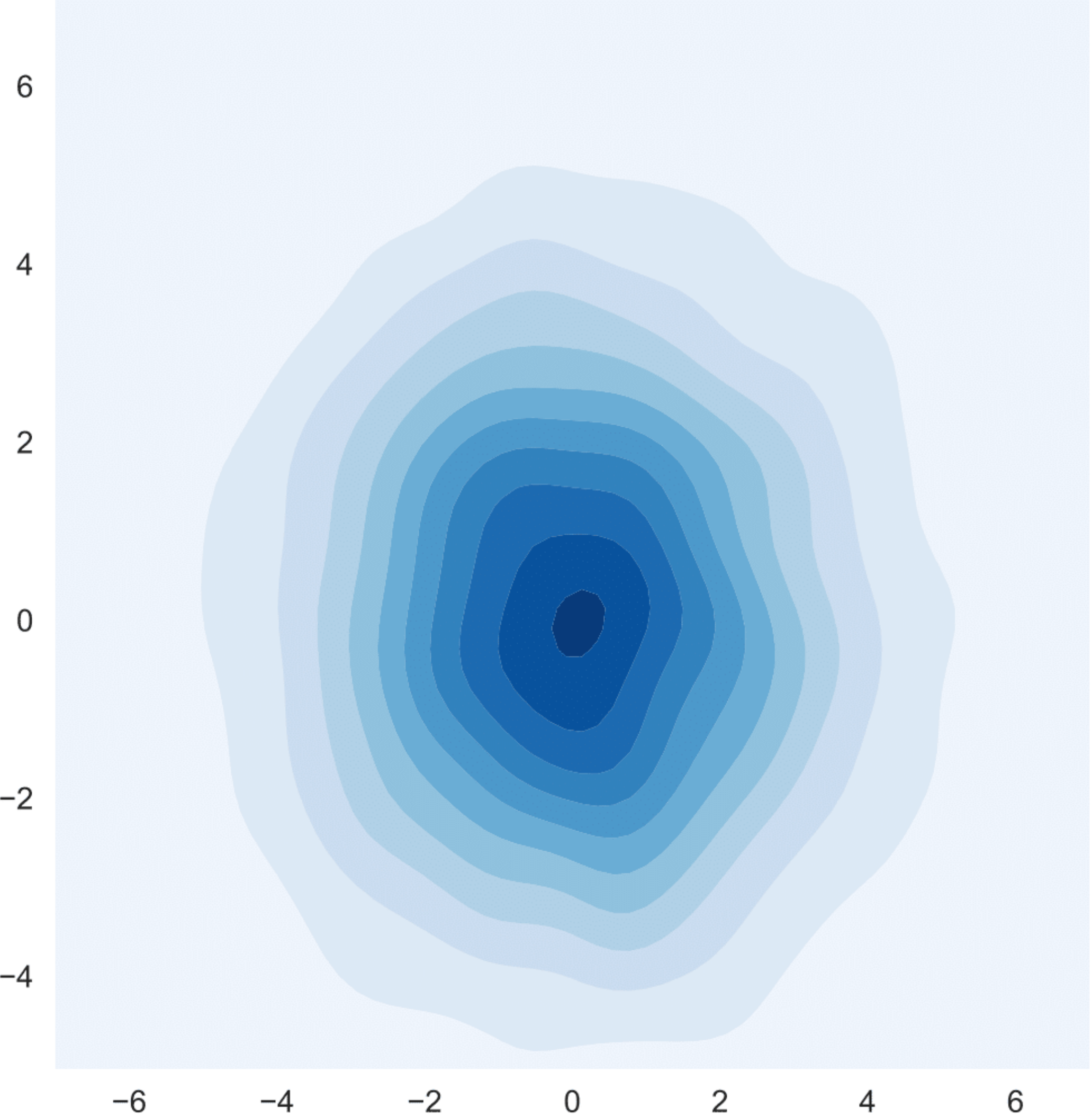} &
\includegraphics[height=0.21\textwidth]{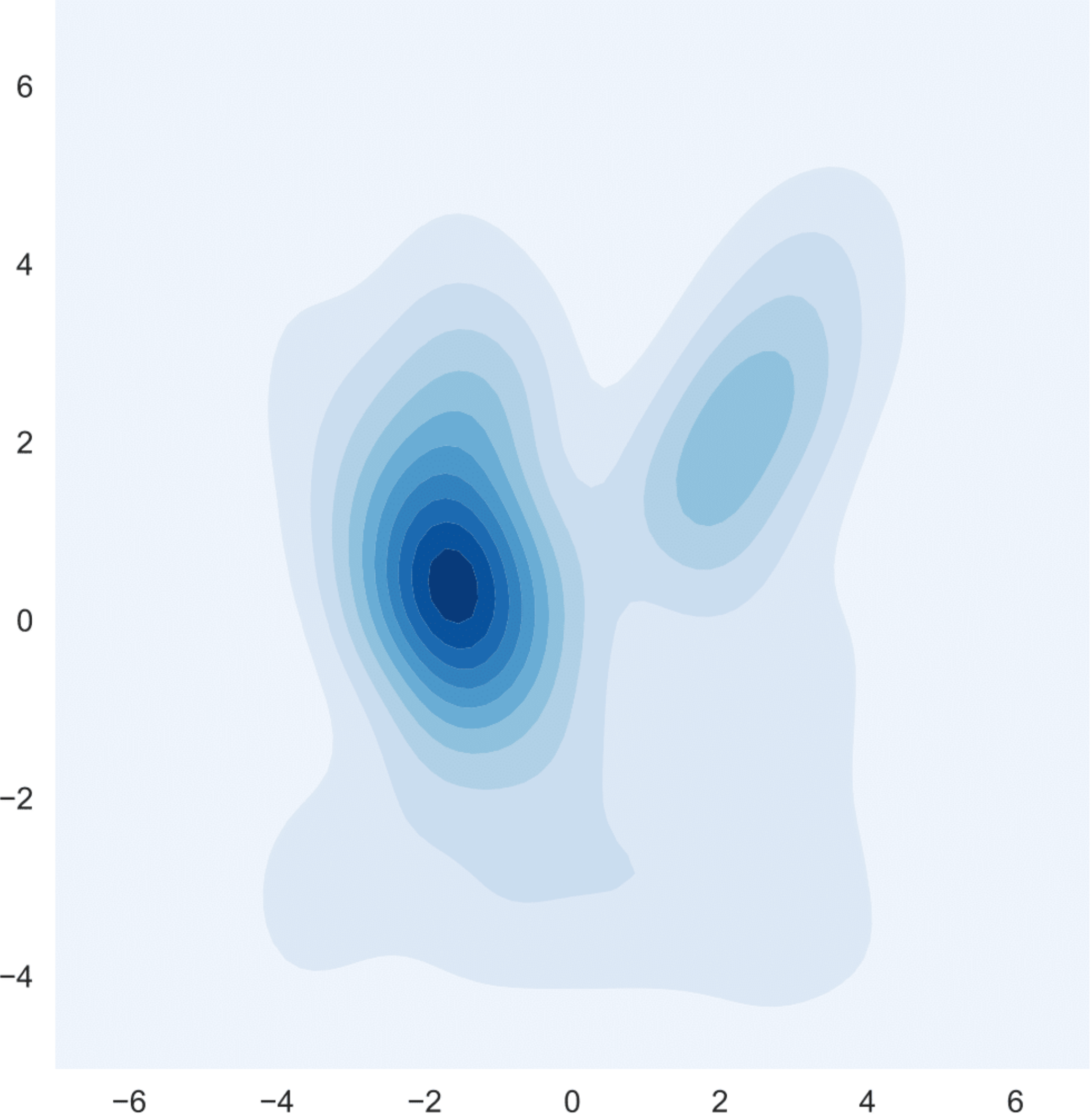} &
\includegraphics[height=0.21\textwidth]{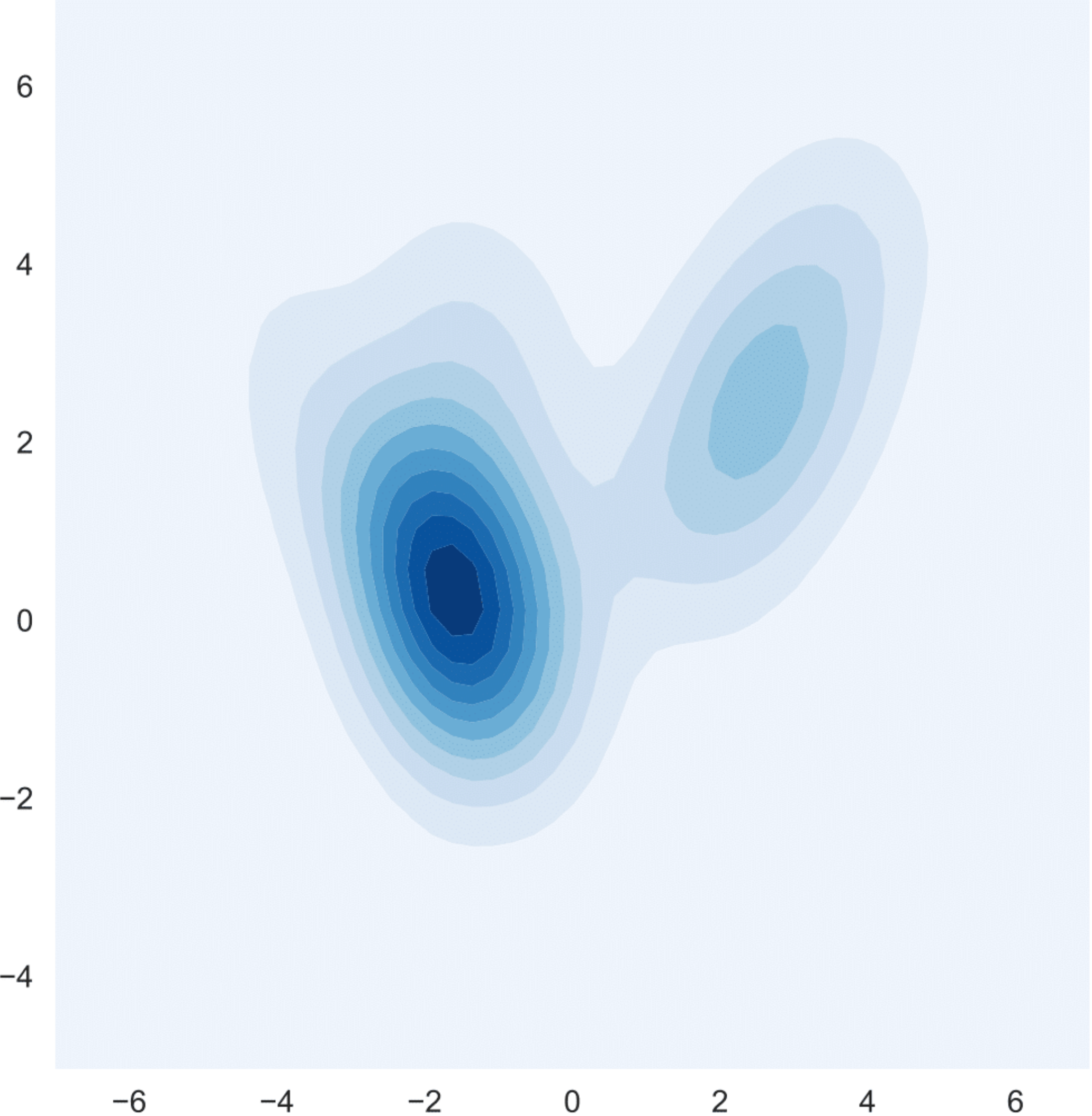} &
\includegraphics[height=0.21\textwidth]{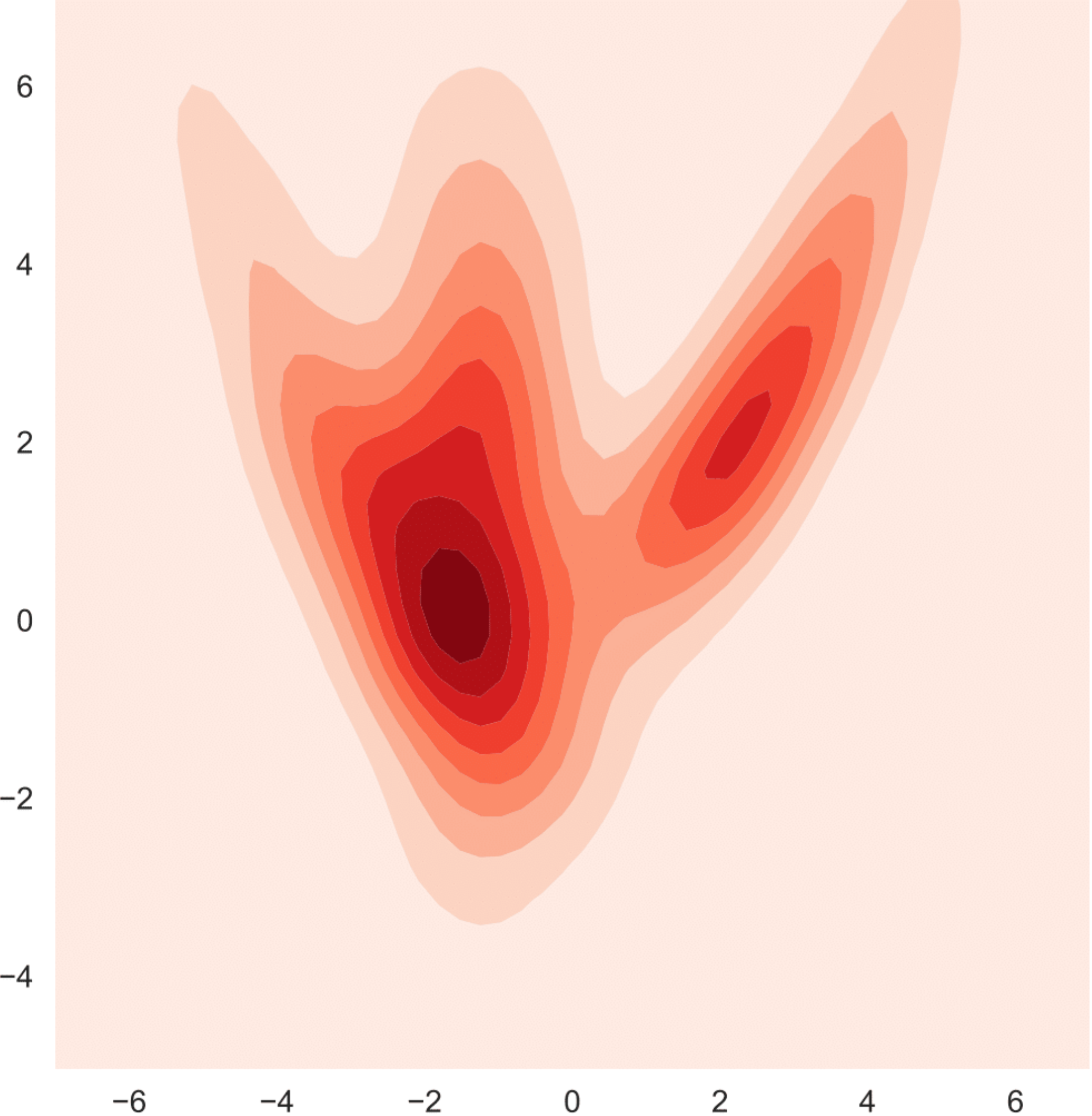} \\
{\small (e) Adap IS, 0 iteration} & {\small (f) Adap IS, 1000 iteration } & {\small (g) Adap IS, 10000 iteration} & {\small (h)  Exact } \\
\end{tabular}
}
\caption{Evolution of the contour of density functions for SteinIS and Adaptive IS. The top row (a)-(d) shows the contours of the evolved density functions in SteinIS, and bottom row (c)-(g) are the evolved contours of the traditional adaptive IS with Gaussian mixture proposals. (h) is the contour of the target density $p$. The number of the mixture components for adaptive IS is $200$ and the number of leader particles for approximating the map in SteinIS is also $200$. }
\label{fig:adap}
\end{figure*}
\subsection{Comparison between SteinIS and Adaptive IS}
In the following, we compare SteinIS with traditional adaptive IS~\citep{ryu2014adaptive} on a probability model $p(\bd{x})$, obtained by applying nonlinear transform on a three-component Gaussian mixture model. 
Specifically, let $ \widetilde{q}$ be a 2D Gaussian mixture model, and 
$\bd T$ is a nonlinear transform defined by $\bd{T}(\bd{z})=[a_1z_1+b_1, a_2 z_1^2+a_3z_2+b_2]^{\top}$, where $\bd{z}=[z_1, z_2]^{\top}.$ We define the target $p$ to be the distribution of $\vx  = \bd T (\bd z)$ when $\bd z \sim \widetilde q$. 
The contour of the target density $p$ we constructed is shown in Figure~\ref{fig:adap}(h). 
We test our SteinIS with $|A| = 100$ particles 
and visualize in Figure~\ref{fig:adap}(a)-(d) the density of the evolved distribution $q_\ell$ using kernel density estimation, by drawing a large number of follower particles. 
We compare our method with 
the adaptive IS by \citep{ryu2014adaptive} 
using a proposal family formed by Gaussian mixture with $200$ components. 
The densities of the proposals obtained by adaptive IS at different iterations
are shown in Figure~\ref{fig:adap}(e)-(g). 
We can see that the evolved proposals of SteinIS converge to the target density $p(\bd{x})$ and approximately match $p(\bd{x})$ at 2000 iterations, but the optimal proposal of adaptive IS with 200 mixture components (at the convergence) can not fit $p(\bd{x})$ well, as indicated by Figure~\ref{fig:adap}(g). This is because the Gaussian mixture proposal family (even with upto 200 components) can not closely approximate the non-Gaussian target distribution we constructed. 
We should remark that SteinIS can be applied to refine 
the optimal proposal given by adaptive IS to get better importance proposal by implementing a set of successive transforms on the given IS proposal. 

Qualitatively, we find that the KL divergence (calculated via kernel density estimation) between our evolved  proposal $q_\ell$ and $p$ decreases to  $\leq 0.003$ after 2000 iterations, while the KL divergence between the optimal adaptive IS proposal  and the target $p$ can be only decreased to $0.42$ even after sufficient optimization. 

\begin{figure}[h]
\centering
\begin{tabular}{cc}
\includegraphics[width=0.2\textwidth]{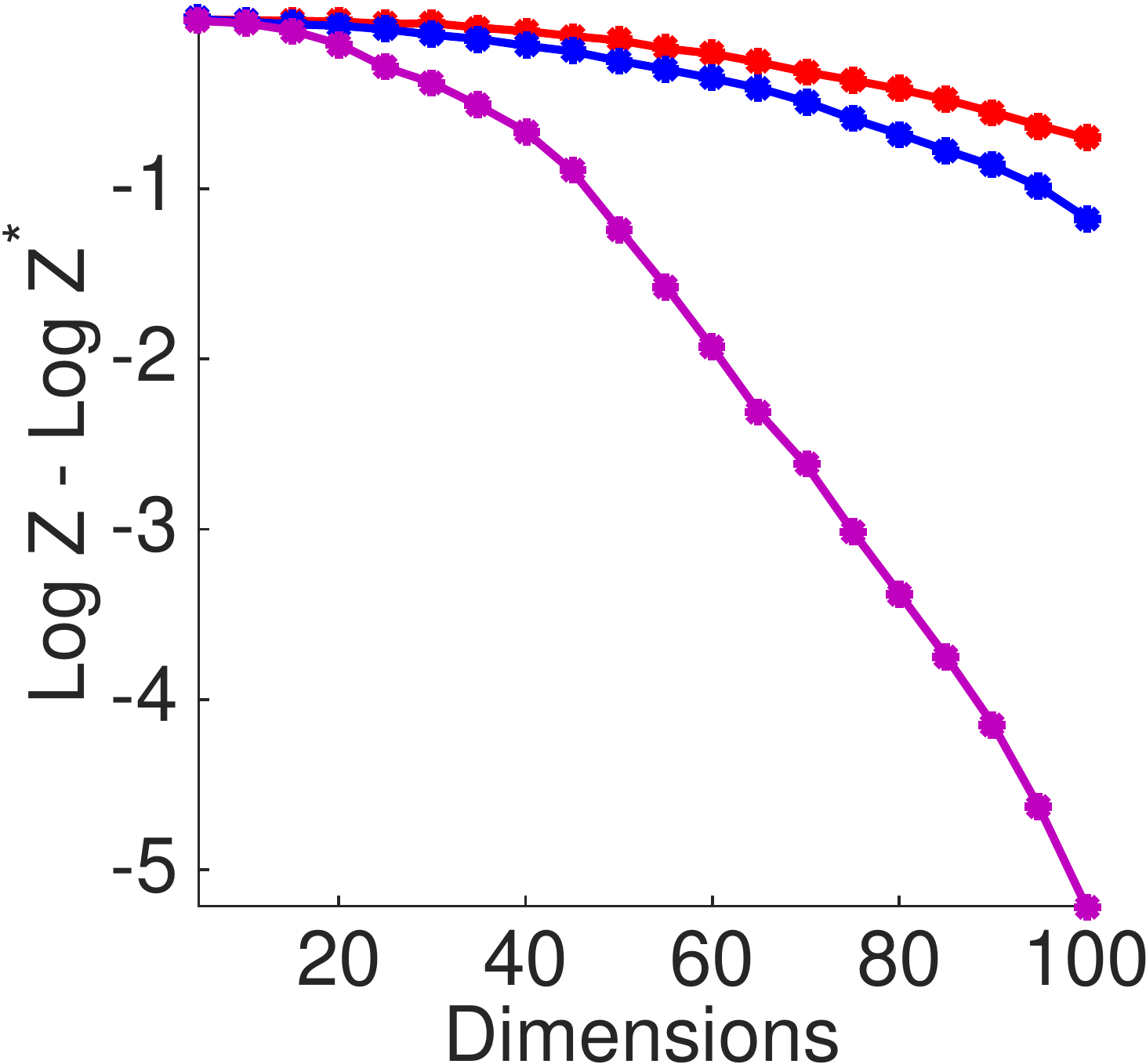} &
\includegraphics[width=0.2\textwidth]{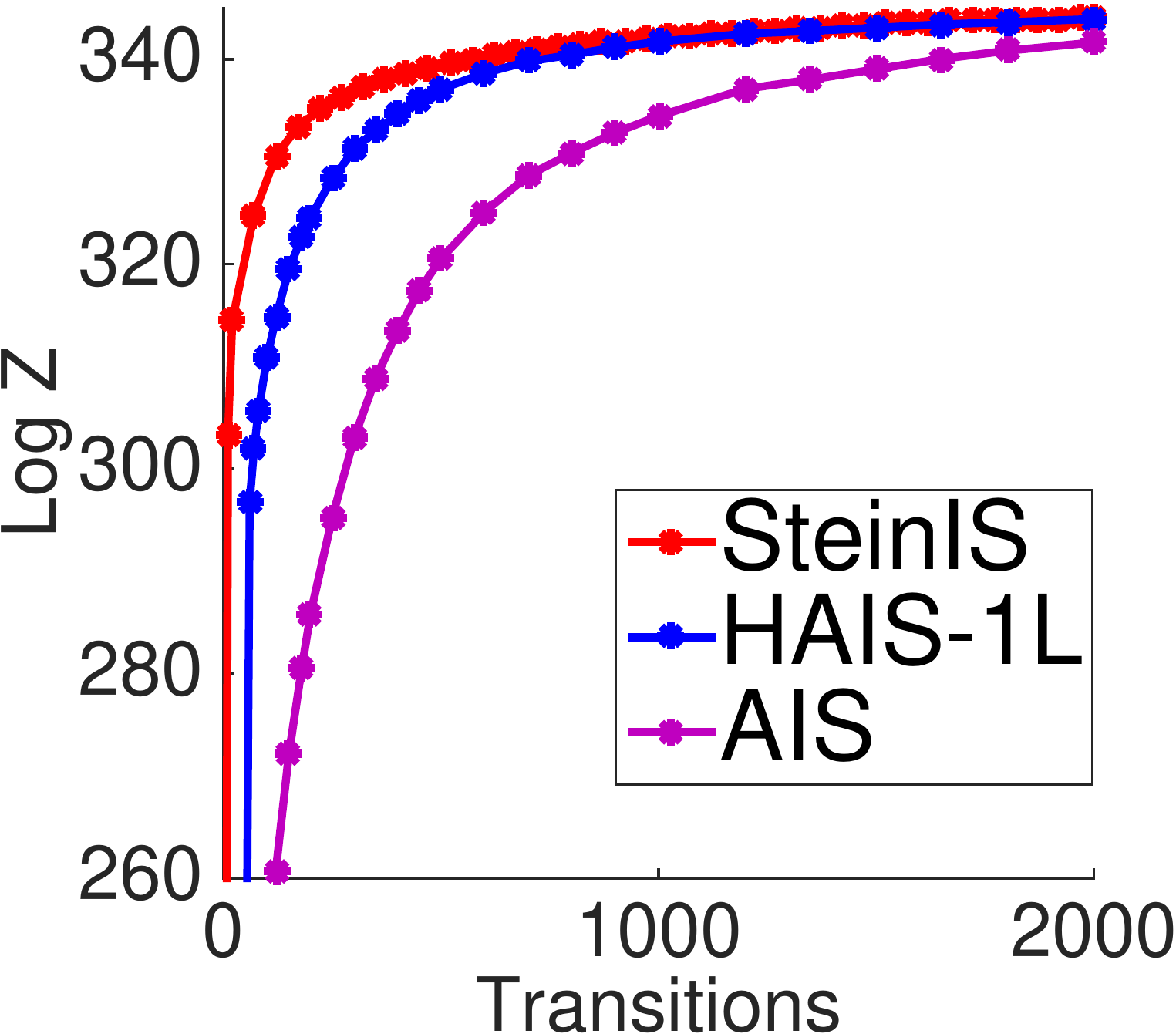} \\
{\small (a)  Vary dimensions} &
{\small (b)  100 dimensions}
\end{tabular}
\caption{Gauss-Bernoulli RBM with $d'=10$ hidden variables. The initial distribution $q_0(\bd{x})$ for all the methods  is a same multivariate Gaussian. We let $|A|=100$ in SteinIS and use $(B=)$100 importance samples in SteinIS, HAIS and AIS. In (a), we use 1500 transitions for HAIS, SteinIS and AIS. "HAIS-1L" means we use $L=1$ leapfrog in each Markov transition of HAIS. 
$\log Z^*$ denotes the logarithm of the exact normalizing constant. All experiments are averaged over 500 independent trails. 
}
\label{fig:GauBerlli}
\end{figure}
\subsection{Gauss-Bernoulli Restricted Boltzmann Machine}
We apply our method to estimate the partition function of Gauss-Bernoulli Restricted Boltzmann Machine (RBM), which is a multi-modal, hidden variable graphical model. It consists of a continuous observable variable $\bd{x}\in \mathbb{R}^d$ and a binary hidden variable $\bd{h}\in \{\pm 1\}^{d'},$ with a joint probability density function of form 
\begin{equation}
p(\bd{x},\bd{h})= \frac{1}{Z} \exp(\bd{x}^\mathrm{T} B\bd{h}+b^\mathrm{T}\bd{x}+c^\mathrm{T}\bd{h}-\frac{1}{2}\| \bd{x}\|_2^2),
\end{equation}
where $p(\bd{x})=\frac{1}{Z}\sum_{\bd{h}} p(\bd{x},\bd{h})$ and $Z$ is the normalization constant. 
By marginalzing the hidden variable $h$, we can show that $p(\bd{x})$ is 
\begin{equation*}
\begin{aligned}
p(\bd{x})=\frac{1}{Z}\exp(b^\mathrm{T}\bd{x}-\frac12\| \bd{x}\|_2^2)\prod_{i=1}^{d'} [\exp(\varphi_i)+ \exp(-\varphi_i)],
\end{aligned}
\end{equation*}
where $\varphi = B^\mathrm{T}\bd{x} + c$, and its score function $\bd{s}_p$ is easily derived as
$$
\bd{s}_p(\bd{x})  = \nabla_{\bd x} \log p(\bd x) = b - \bd{x} + B\frac{\exp(2\varphi)-1}{\exp(2\varphi)+1}.
$$

In our experiments, we simulate a true model $p(\bd{x})$ by drawing $b$ and $c$ from standard Gaussian and select $B$ uniformly random from $\{0.5, -0.5\}$ with probability 0.5. The dimension of the latent variable $\bd{h}$ is 10 so that the probability model $p(\bd{x})$ is the mixture of  $2^{10}$ multivariate Gaussian distribution. The exact normalization constant $Z$ can be feasibly calculated using the brute-force algorithm in this case. 
Figure~\ref{fig:GauBerlli}(a) and Figure \ref{fig:GauBerlli}(b) shows the performance of SteinIS on Gauss-Bernoulli RBM 
when we vary the dimensions of the observed variables 
and the number of transitions in SteinIS, respectively.  
 We can see that SteinIS converges slightly faster than HAIS which uses one leapfrog step in each of its Markov transition. Even with the same number of Markov transitions, AIS with Langevin dynamics converges much slower than both SteinIS and HAIS. 
The better performance of HAIS comparing to AIS was also observed by ~\citet{sohl2012hamiltonian} when they first proposed HAIS. 

\begin{figure}[t]
\centering
\begin{tabular}{cc}
\includegraphics[width=0.2\textwidth]{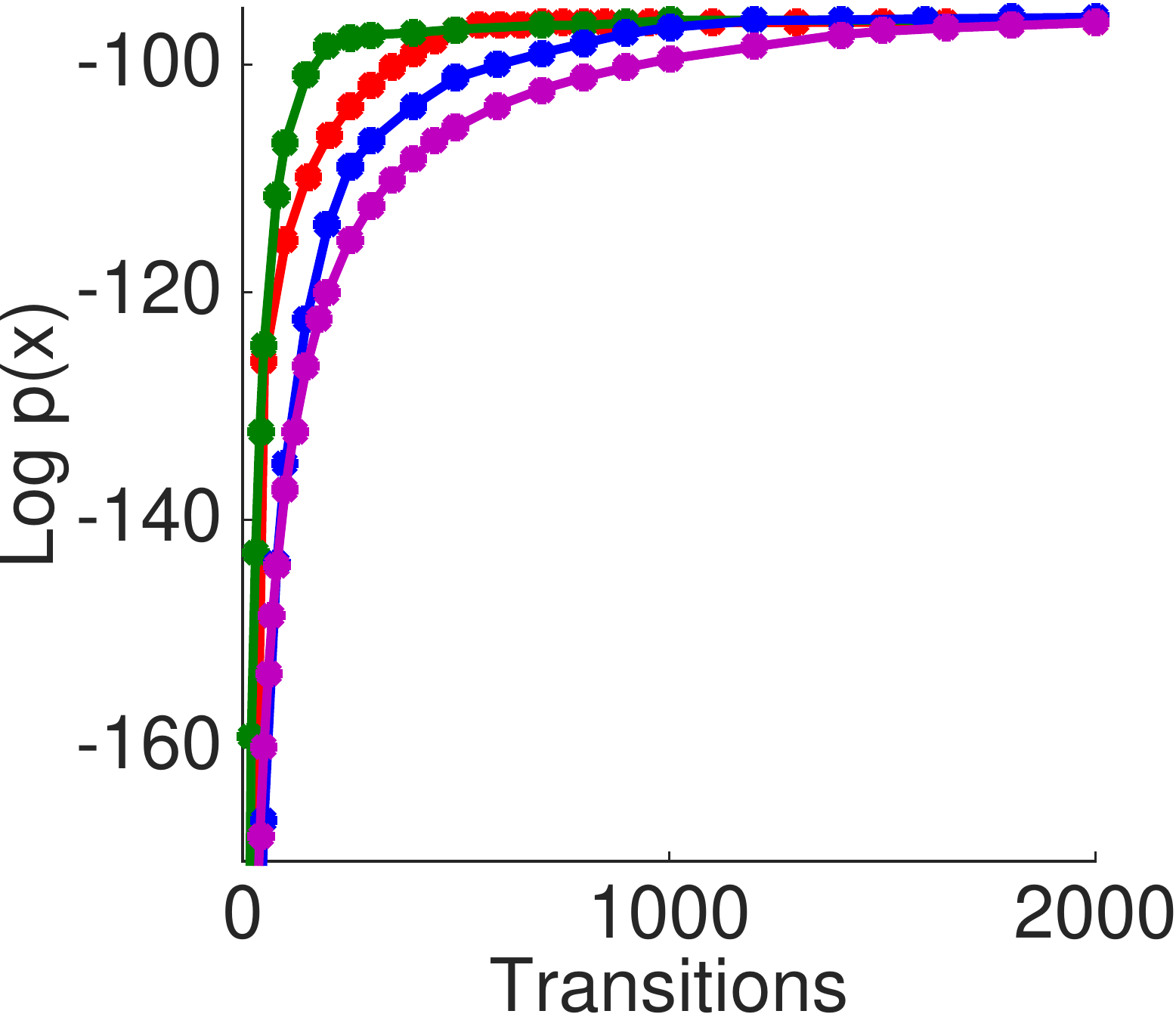}&
\includegraphics[width=0.2\textwidth]{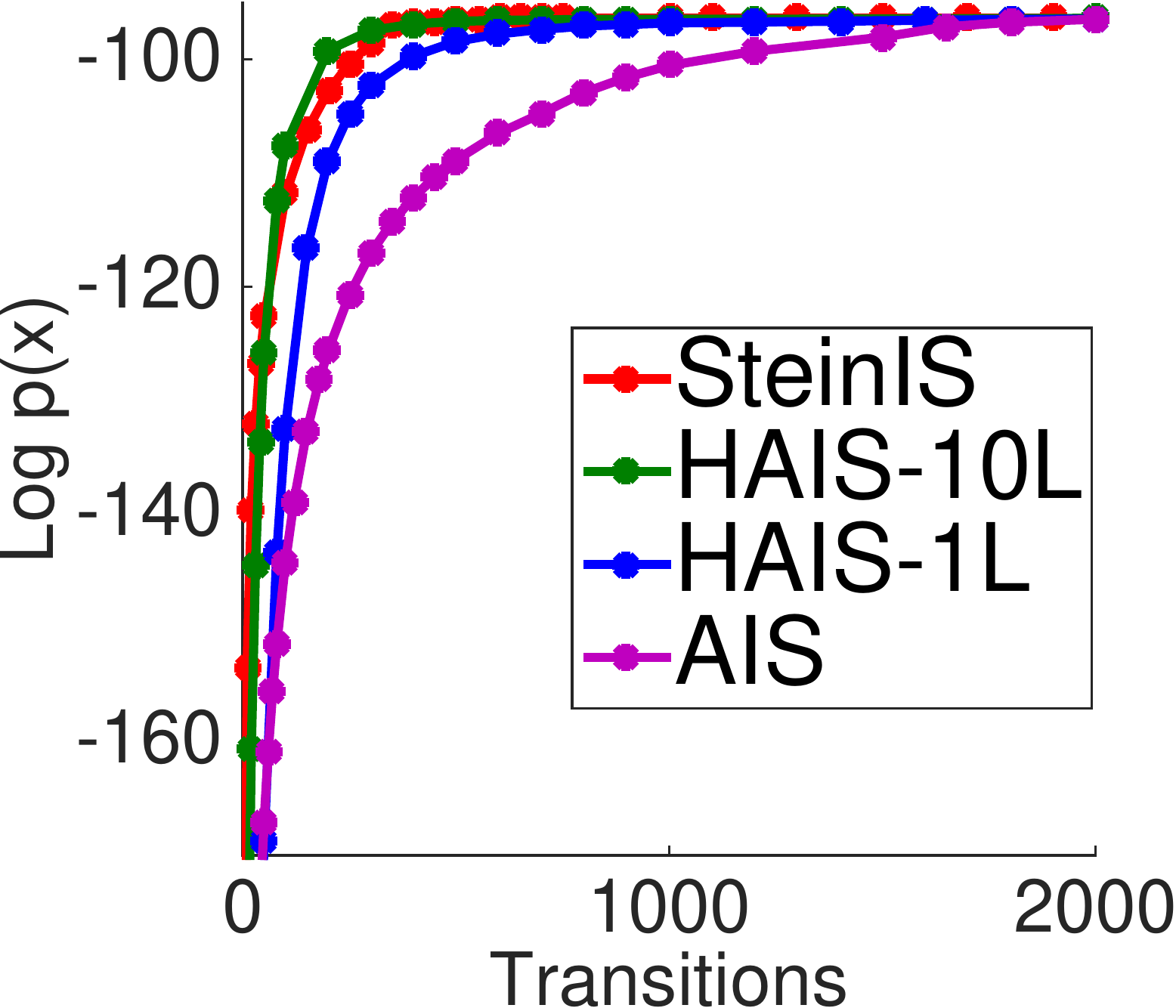} \\
{\small (a)  20 hidden variables}&
{\small (b)  50 hidden variables}
\end{tabular}
\caption{
Calculating the testing log-likelihood $\log p(x)$ for the deep generative model on MNIST. 
The initial distribution $q_0$ used in SteinIS, HAIS and AIS is a same multivariate Gaussian. 
We let $|A|=60$ in SteinIS and use 60 samples for each image to implement IS in HAIS and AIS. "HAIS-10L" and "HAIS-1L" denote using $L=10$ and $L=1$ in each Markov transition of HAIS, respectively. The log-likelihood $\log p(x)$ is averaged over 1000 images randomly chosen from MNIST. 
Figure (a) and (b) show the results when using $20$ and $50$ hidden variables, respectively. 
Note that the dimension of the observable variable $\vx$ is fixed, and is the size of the MNIS images. }
\label{figdecode}
\end{figure}
\subsection{Deep Generative Models}
 Finally, we implement our SteinIS to evaluate the $\log$-likelihoods of the decoder models in variational autoencoder (VAE) \citep{kingma2013auto}. VAE is a directed probabilistic graphical model. The decoder-based generative model is defined by a joint distribution over a set of latent random variables $\bd{z}$ and the observed variables $\bd{x}: p(\bd{x}, \bd{z})=p(\bd{x}\mid \bd{z}) p(\bd{z}).$ We use the same network structure as that in \citet{kingma2013auto}. The prior $p(\bd{z})$ is chosen to be a multivariate Gaussian distribution. The log-likelihood is defined as $p(\bd{x})=\int p(\bd{x}\mid\bd{z})p(\bd{z})d\bd{z},$ where $p(\bd{x}\mid \bd{z})$ is the Bernoulli MLP as the decoder model given in \citet{kingma2013auto}.
 In our experiment, we use a two-layer network for $p(\bd{x}\mid\bd{z}),$ whose parameters are estimated using a standard VAE based on the MNIST training set.  
 For a given observed test image $\vx$, we use our method to sample the posterior distribution $p(\bd z | \vx)  = \frac{1}{p(\vx)}  p(\bd{x} | \bd{z}) p(\bd{z})$, and estimate the partition function $p(x)$, which is the testing likelihood of image $\vx$. 

Figure~\ref{figdecode} also indicates that our SteinIS converges slightly faster than HAIS-1L which uses one leapfrog step in each of its Markov transitions, denoted by HAIS-1L. Meanwhile, the running time of SteinIS and HAIS-1L is also comparable as provided by Table~\ref{timetable}. Although HAIS-10L, which use 10 leapfrog steps in each of its Markov transition, converges faster than our SteinIS, it takes much more time than our SteinIS in our implementation since the leapfrog steps in the Markov transitions of HAIS are sequential and can not be parallelized. See Table~\ref{tab:tab1}.  
Compared with HAIS and AIS, our SteinIS has another advantage: if we want to increase the transitions from 1000 to 2000 for better accuracy, SteinIS can build on the result from 1000 transitions and just need to run another 1000 iterations, while HAIS cannot take advantage of the result from 1000 transitions and have to independently run another 2000 transitions. 

\begin{table}[h]
\label{tab:tab1}
\centering
\caption{Running Time (in seconds) on  MNIST, using the same setting as in Figure~\ref{figdecode}. We use 1000 transitions in all methods to test the running time.}
\vspace{3bp} \begin{tabular}{|c|c|c|c|}
  \hline
  Dimensions of $\bd{z}$ & 10  & 20  & 50  \\ \hline
  SteinIS & 224.40 & 226.17 & 261.76 \\ \hline
  HAIS-10L & 600.15 & 691.86 & 755.44 \\ \hline
  HAIS-1L & 157.76 & 223.30 & 256.23 \\ \hline
  AIS & 146.75 & 206.89 & 230.14 \\
  \hline
\end{tabular}
\label{timetable}
\end{table}

\section{CONCLUSIONS}
In this paper, we propose an nonparametric adaptive importance sampling algorithm which leverages the nonparametric transforms of SVGD to maximumly decrease the KL divergence between our importance proposals and the target distribution. Our algorithm turns SVGD into a typical adaptive IS for more general inference tasks. Numerical experiments demonstrate that our SteinIS works efficiently on the applications such as estimating the partition functions of graphical models and evaluating the log-likelihood of deep generative models. 
Future research includes 
improving the computational and statistical efficiency in high dimensional cases, 
more theoretical investigation on the convergence of $\KL(q_\ell ~||~p)$, and incorporating Hamiltonian Monte Carlo into our SteinIS to derive more efficient algorithms.

{\bf Acknowledgments}\\ 
This work is supported in part by NSF CRII 1565796. We thank Yuhuai Wu from Toronto for his valuable comments.

\bibliographystyle{icml2017}
\bibliography{main.bib}

\end{document}